\documentclass{article}

\PassOptionsToPackage{numbers}{natbib}

\usepackage[final]{neurips_2023}

\usepackage{amsfonts}
\usepackage{amsmath}
\usepackage{amsthm}
\usepackage{amssymb}
\usepackage{dsfont}

\usepackage{xcolor}
\usepackage{color}
\usepackage{graphicx}

\usepackage{verbatim}
\usepackage{xspace} %

\usepackage{enumerate}
\usepackage{enumitem}

\providecommand{\norm}[1]{\left\lVert#1\right\rVert}

\providecommand{\R}{\mathbb{R}} %

\DeclareMathOperator{\E}{{\mathbb E}}

\providecommand{\0}{\mathbf{0}}
\providecommand{\1}{\mathbf{1}}
\renewcommand{\aa}{\mathbf{a}}
\providecommand{\bb}{\mathbf{b}}

\renewcommand{\gg}{\mathbf{g}}

\providecommand{\vv}{\mathbf{v}}

\providecommand{\xx}{\mathbf{x}}
\providecommand{\yy}{\mathbf{y}}
\providecommand{\zz}{\mathbf{z}}

\providecommand{\mA}{\mathbf{A}}
\providecommand{\mB}{\mathbf{B}}
\providecommand{\mC}{\mathbf{C}}
\providecommand{\mD}{\mathbf{D}}

\providecommand{\mG}{\mathbf{G}}
\providecommand{\mH}{\mathbf{H}}
\providecommand{\mI}{\mathbf{I}}

\providecommand{\mS}{\mathbf{S}}

\providecommand{\mU}{\mathbf{U}}
\providecommand{\mV}{\mathbf{V}}

\providecommand{\mX}{\mathbf{X}}

\providecommand{\mZ}{\mathbf{Z}}

\providecommand{\cA}{\mathcal{A}}

\providecommand{\cN}{\mathcal{N}}
\providecommand{\cO}{\mathcal{O}}

\providecommand{\cT}{\mathcal{T}}

\usepackage{bm}

\providecommand{\mLambda}{\boldsymbol{\Lambda}}
\providecommand{\mlambda}{\boldsymbol{\lambda}}

\newcommand{\dpsgd}{DP-SGD\xspace}
\newcommand{\pgd}{PGD\xspace}
\newcommand{\antipgd}{Anti-PGD\xspace}
\newcommand{\chesspgd}{Chess-PGD\xspace}
\newcommand{\dpftrl}{DP-FTRL\xspace}
\newcommand{\dpmf}{MF-DP-FTRL\xspace}
\DeclareMathOperator{\chess}{\text{chess}}

\usepackage{url}

\renewcommand{\epsilon}{\varepsilon}

\DeclareMathOperator{\sens}{\operatorname{sens}}

\usepackage[utf8]{inputenc} %
\usepackage[T1]{fontenc}    %
\usepackage{hyperref}       %
\usepackage{url}            %
\usepackage{booktabs}       %
\usepackage{amsfonts}       %
\usepackage{nicefrac}       %
\usepackage{microtype}      %
\usepackage{xcolor}         %
\usepackage{adjustbox}
\usepackage{graphicx}
\usepackage{caption}
\usepackage{subfigure}

\usepackage{amsmath}
\usepackage{amssymb}
\usepackage{mathtools}
\usepackage{amsthm}

\usepackage[capitalize]{cleveref}

\theoremstyle{plain}
\newtheorem{theorem}{Theorem}[section]
\newtheorem{proposition}[theorem]{Proposition}
\newtheorem{lemma}[theorem]{Lemma}

\theoremstyle{definition}

\newtheorem{assumption}[theorem]{Assumption}
\newtheorem{example}[theorem]{Example}
\theoremstyle{remark}

\newtheorem{problem}[theorem]{Problem}

\usepackage[textsize=tiny]{todonotes}

\usepackage{fancyhdr}

\usepackage{xcolor}

\title{Gradient Descent with Linearly Correlated Noise: Theory and Applications to Differential Privacy}

\author{%
  Anastasia Koloskova\thanks{Work performed while doing an internship at Google Research. Correspondence to: Anastasia Koloskova <anastasia.koloskova@epfl.ch>, Ryan McKenna
<mckennar@google.com>.} \\
  EPFL, Switzerland
  \And
  Ryan McKenna\\
  Google Research
  \And
  Zachary Charles \\
  Google Research
  \And
  Keith Rush\\
  Google Research
  \And
  Brendan McMahan\\
  Google Research
}

\begin{document}

\maketitle

\begin{abstract}
We study gradient descent under linearly correlated noise.
Our work is motivated by recent practical methods for optimization with differential privacy (DP), such as \dpftrl, which achieve strong performance in settings where privacy amplification techniques are infeasible (such as in federated learning).
These methods inject privacy noise through a matrix factorization mechanism, making the noise linearly correlated over iterations. 
We propose a simplified setting that distills key facets of these methods and isolates the impact of linearly correlated noise. We analyze the behavior of gradient descent in this setting, for both convex and non-convex functions.
Our analysis is demonstrably tighter than prior work and recovers multiple important special cases exactly (including anti-correlated perturbed gradient descent).
We use our results to develop new, effective matrix factorizations for differentially private optimization, and highlight the benefits of these factorizations theoretically and empirically.

\end{abstract}

\section{Introduction}

Differential privacy (DP) is a critical framework for designing algorithms with provable statistical privacy guarantees. DP stochastic gradient descent (\dpsgd,~\citet{dpsgd_2016}) is particularly important for enabling private empirical risk minimization (ERM) of machine learning models. Many works have analyzed the convergence behavior of DP ERM methods, including \dpsgd~\citep{bassily2019private, feldman2022hiding, wang2017differentially, das2022beyond}. However, obtaining good privacy/utility trade-offs with \dpsgd can require excessively large batch sizes or privacy amplification techniques such as subsampling~\citep{bassily2014private, bassily2019private, zhu2019poission} and shuffling~\citep{erlingsson2019amplification, feldman2022hiding}. In some applications, including cross-device federated learning, limited and device-controlled client availability can make sampling or shuffling infeasible~\citep{kairouz2021advances}. Even outside of such applications, many implementations of \dpsgd do not properly use the Poisson subsampling scheme analyzed by \citet{dpsgd_2016} for amplification, and instead use a single fixed permutation of the dataset~\citep{choquette22:multi-epochs}.

\citet{kairouz21-dp-ftrl} propose an alternative method, \dpftrl, which can attain good privacy/utility trade-offs without amplification. Their key insight is that for SGD-style algorithms, the variance on \emph{prefix sums} $\gg_0 + \dots + \gg_t$, $t \in \{1, \dots, T\}$ of gradients $\gg_j$ is more important than the variance on individual gradients.
By adding carefully tailored noise that is \emph{linearly correlated} over iterations to the gradients, one can reduce the error on the prefix sums, at the cost of increased error on the individual gradients, for a fixed privacy budget. The \dpftrl mechanism is competitive with or better than \dpsgd, even without relying on privacy amplification, and enabled \citet{mcmahan2022federated} to train the first differentially private machine learning model on user data in a production setting. 

\begin{figure}[t]
    \centering
    \includegraphics[width=0.8\linewidth]{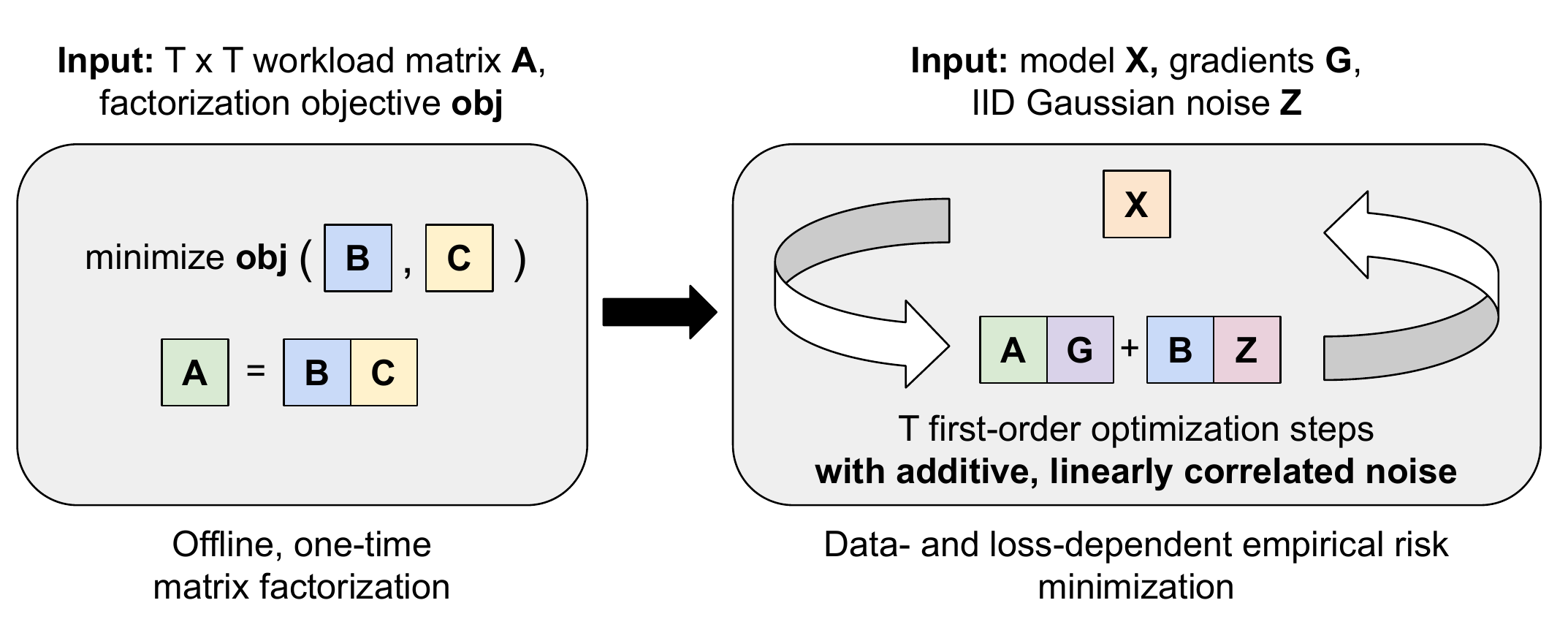}
    \caption{Two-stage \dpmf workflow proposed by \citet{denisov2022:matrix-fact}. The user selects a workload matrix $\mA$ representing a desired first-order optimization method. Offline, the user finds a factorization $\mB\mC = \mA$, using an objective that balances ERM performance (as a function of $\mB$) and privacy (as a function of $\mC$). The user applies $\mA$ to a downstream ERM task, but with linearly correlated additive noise governed by $\mB$.}
    \label{fig:mf_workflow}
\end{figure}

\citet{denisov2022:matrix-fact,choquette22:multi-epochs} develop a refinement of \dpftrl, \dpmf, by formulating and solving an offline matrix factorization problem to find the ``optimal'' correlated noise structure under DP constraints. That is, for a fixed privacy level, they aim to find correlated noise structures that lead to improved optimization. A simplified diagram of their workflow is given in \cref{fig:mf_workflow}.
However, (as we detail in \cref{sec:matrix-factorization}) their offline factorization objective is based on an online convergence bound that is loose. This raises questions about whether there are factorization objectives that better capture convergence behavior of gradient descent algorithms with correlated noise.

In this paper we study this class of mechanisms more closely and provide a detailed analysis of linearly correlated noise from an optimization point of view. Our main contributions are as follows:
\begin{itemize}
    \item We propose a novel stochastic optimization problem that extracts key facets of methods like (MF-)DP-FTRL, and which isolates the effects of linearly correlated noise on optimization.
    \item We derive convergence rates for gradient descent on smooth convex and non-convex functions in such settings that showcase the effect of linearly correlated noise and recover tight convergence rates in notable special cases. We use a novel proof technique that may be of independent interest.
    \item We use this theory to design a new objective for the offline matrix factorization workflow in \cref{fig:mf_workflow}. We show that solving this objective leads to \dpmf mechanisms with improved convergence properties. We validate the mechanism empirically on a variety of datasets and tasks, matching or outperforming prior methods.
\end{itemize}

\subsection{Related Work}
\paragraph{Matrix mechanisms for differential privacy.}
Our work is closely related to  differentially private optimization using matrix mechanisms~\citep{Li2015TheMM}. Historically, such mechanisms were applied to linear statistical queries~\citep{li2010optimizing, mckenna2018optimizing, edmonds2020power, constant_matters}. \citet{denisov2022:matrix-fact} and \citet{choquette22:multi-epochs} extended these mechanisms to the adaptive streaming setting, allowing their application to optimization with DP. \citet{denisov2022:matrix-fact} show that this framework (\dpmf) subsumes and improves the \dpftrl algorithm~\citep{kairouz21-dp-ftrl}. Both \dpftrl and \dpmf improve privacy guarantees relative to \dpsgd~\citep{dpsgd_2016} without amplification, and can be combined with techniques such as momentum for improved utility~\cite{tran2022momentumFTRL}. The aforementioned work focuses on methods for computing factorizations, privacy properties, and empirics. Our work studies the analytic relationship between the correlated noise induced by the \dpmf framework and the downstream effect on optimization performance. 

\paragraph{SGD with correlated noise.}
Stochastic noise in optimization arises in a variety of ways, including mini-batching~\citep{Dekel12:sgd_convex_proof} and explicit noise injection~\citep{duchi2012randomized, zhou2019toward, jin2021nonconvex}.
While most analyses of SGD assume this noise is independent across iterates, some work considers correlated noise. For example, shuffle SGD involves correlated noise due to sampling without replacement \citep{mishchenko2020random, yun2022minibatch}. %
\citet{lucchi22:noise_correlation} use correlated Brownian motion to improve SGD's ability to explore the loss landscape. Recently, \citet{orvieto2022anticorrelated,orvieto2022explicit} investigated anti-correlated noise as a way to impose regularization and improve generalization. We consider a linearly correlated noise model,
and analyze its impact on SGD's convergence to critical points.

\paragraph{SGD with biased noise.} Many algorithms can be viewed as SGD with structured but potentially biased noise, including SGD with (biased) compression~\citep{stich2018sparsified, gorbunov2020linearly}, delayed SGD~\citep{mania17:perturbed_analysis, dutta2018slow}, local SGD~\citep{stich2018local}, federated learning methods~\citep{karimireddy2020scaffold, yuan2020federated, mitra2021linear, nguyen2022federated}, decentralized optimization methods \cite{yu2019linear, koloskova20:unified}, and many others. Convergence analyses for such methods often use techniques like perturbed iterate analysis~\citep{mania17:perturbed_analysis}.  Correlated gradient noise also biases the gradient updates. However, as we show in Section~\ref{sec:theory}, directly applying such techniques to linearly correlated noise does not lead to tight convergence guarantees.

\section{Background}\label{sec:matrix-factorization}

In this work, we focus on an empirical risk minimization (ERM) problem of the form
\begin{equation}\label{eq:erm}
    \min_{\xx \in \R^d} \left[f(\xx) = \frac{1}{n} \sum_{i = 1}^n l(\xx, \xi_i)\right],
\end{equation}
where $l(\xx, \xi_i)$ is the loss of a model $\xx$ on a data point $\xi_i$, and $n$ is the training set size. We would like to solve \eqref{eq:erm} while guaranteeing some form of privacy for the training set. We focus on \emph{differential privacy} (DP,~\citep{dp_def}), a widely-used standard for anonymous data release. DP guarantees statistical difficulty in distinguishing whether or not a particular unit's data served as an input to a given algorithm, based on the algorithm's output. This protected unit may represent a single training example or a semantically higher-level unit like the entirety of a user's data.

While there are many methods for solving \eqref{eq:erm}, we will follow \citet{denisov2022:matrix-fact,choquette22:multi-epochs} and restrict to first-order algorithms $\mathfrak{A}$ that linearly combine (stochastic) gradients. Each algorithm $\cA \in \mathfrak{A}$ is parameterized by a learning rate $\gamma > 0$, a number of steps $T > 0$, and scalars $\{a_{tj}\}_{1 \leq j \leq t \leq T}$. Given a starting point $\xx_0$, $\mathcal{A}$ produces iterates $\xx_t \in \R^d$ given by
\begin{align*}
    \xx_{t + 1} = \xx_0 -\gamma\cA_t(\gg_1, \dots, \gg_t) && \cA_t(\gg_1,\dots,\gg_t) =\textstyle \sum_{j = 1}^t a_{tj} \gg_j
\end{align*}
where $\gg_t$ is a (mini-batch) gradient of $f$ computed at $\xx_t$.
This class encompasses a variety of first-order algorithms, including SGD \cite{Robbins51:sgd_original}, SGD with momentum \cite{polyak1964:momentum,nesterov1983:momentum}, and delayed SGD~\citep{Agarwal11:delayedSGD}. This class also captures algorithms that use learning rate scheduling, so long as the schedule is independent of the gradient values.
We re-write the output of $\cA$ in matrix notation by defining:
\begin{align*}
    \mX &= \left[\xx_1, \dots, \xx_T\right]^\top \in \R^{T \times d},~\mX_0 = \left[\xx_0, \dots, \xx_0\right]^\top \in \R^{T \times d}\\
    \mG &= \left[\gg_1, \dots, \gg_T\right]^\top \in \R^{T \times d},~
    \mA = \left[a_{ij}\right]_{1 \leq i,j \leq T} \in \R^{T\times T}
\end{align*}
Here $\mA$ is the \emph{workload matrix} representing $\mathcal{A}$. At iteration $t$, $\mathcal{A}$ can only use the current and previous gradients, so $a_{tj} = 0$ for $j > t$ (ie. $\mA$ is lower-triangular).
In this notation, the iterates of $\mathcal{A}$ satisfy
\begin{align}\label{eq:algo}
    \mX = \mX_0 - \gamma\mA \mG.
\end{align}

\begin{example}[SGD]\label{ex:prefix-sum}
Define the \emph{prefix-sum} matrix $\mS \in \R^{T\times T}$ as the all-ones lower-triangular matrix.
If $\mA = \mS$, then \eqref{eq:algo} is simply SGD with learning rate $\gamma$. As discussed by \citet[Section 4]{denisov2022:matrix-fact}, we also recover SGD with momentum using an appropriate transformation $\mS'$ of $\mS$.
\end{example}

\subsection{Matrix Factorization and Privacy Mechanisms}
In order to make the output of \eqref{eq:algo} differentially private, we typically need to clip the gradients and add noise. Let $\overline\mG$ denote the matrix whose rows (gradients) have been clipped to some $\ell_2$ threshold $\alpha$. Let $\mZ \in \R^{T\times d}$ be a matrix with entries drawn independently from $\cN(0, \nicefrac{\zeta^2}{d})$.
The well-known \dpsgd algorithm~\citep{dpsgd_2016} adds this noise to each clipped gradient, so that
\begin{equation}\label{eq:dp_sgd}
\mX = \mX_0 -\gamma \mA(\overline\mG + \mZ).
\end{equation}
For consistency, we consider \eqref{eq:algo} to be the special case of \eqref{eq:dp_sgd} where $\mZ = \0$ and $\alpha = \infty$.
The variance $\zeta^2$ depends on the clipping threshold $\alpha$ and desired $(\epsilon, \delta)$ privacy we aim to achieve \citep{dpsgd_2016}. 

To derive algorithms with improved DP guarantees, \citet{denisov2022:matrix-fact} add the noise $\mZ$ to a factorized version of $\mA$. For a factorization $\mA = \mB \mC$ with $\mB, \mC \in \R^{T \times T}$,
we add noise to the iterates via:
\begin{align}\label{eq:dp-algo}
    \mX = \mX_0 -\gamma \mB \left(\mC \overline\mG + \sens(\mC)\mZ \right) \equiv \mX_0 -\gamma \left(\mA\overline\mG + \sens(\mC)\mB \mZ\right). %
\end{align}
Here, $\operatorname{sens}(\mC)$ is a number representing the sensitivity of the mapping $\overline\mG \mapsto \mC \overline\mG$ to ``adjacent'' input changes. We note that the sensitivity changes depending on the notion of adjacency. In single-epoch settings, two input matrices are adjacent if they differ by a single row~\citep{denisov2022:matrix-fact}, so the sensitivity function is $\operatorname{sens}(\mC) := \max_{i \in \{1, \dots, T\}}\|\mC_{[:, i]}\|_2$, i.e. the maximum $\ell_2$-squared column norm of $\mC$.
For details and extensions to multiple epochs, see \citep{choquette22:multi-epochs}.

If the variance of entries of $\mZ$ is fixed to some value $\nicefrac{\zeta^2}{d}$, then for all the possible factorizations $\mA = \mB \mC$ in \eqref{eq:dp-algo} have exactly same privacy guarantees, depending only on $\zeta$. It will also be convenient to define $\sigma = \sens(\mC) \zeta$ as the 'effective' variance of $\mZ$ after re-scaling by the sensitivity. Note that for a fixed $\sigma$, the privacy guarantees of \eqref{eq:dp-algo} might be different depending on the sensitivity. 

The factorization $\mB = \mA, \mC = \mI$ recovers \dpsgd \eqref{eq:dp_sgd}, but factorizations with better privacy-utility trade-offs may exist.
The formulation of~\cref{eq:dp-algo} transfers the linear optimization algorithm~\eqref{eq:algo} into the setting of the matrix mechanism~\citep{Li2015TheMM}, a well-studied family of mechanisms in differential privacy.
\citet{denisov2022:matrix-fact,choquette22:multi-epochs} show that the mechanism in \cref{eq:dp-algo} provides a DP guarantee equivalent to a single application of the Gaussian mechanism, which can be computed tightly using numerical accounting techniques~\citep{wang2019subsampled, koskela2021tight}.

\paragraph{Finding good factorizations.} 
Intuitively, a factorization $\mA = \mB\mC$ is good if $\sens(\mC)$ is small and the added noise $\mB \mZ$ does not significantly degrade the convergence of \eqref{eq:dp-algo}. In order to quantify the effect of this added correlated noise on optimization, \citet{denisov2022:matrix-fact} derive an online regret bound for \eqref{eq:dp-algo} in the convex case against an adaptive adversary. Translating this via online-to-batch convergence to the stochastic setting, the iterates $\xx_t$ satisfy
\begin{equation}\label{eq:prior_rate}
    \frac{1}{T + 1} \sum_{t = 0}^T \E \left[f(\xx_t) - f^\star\right] \leq \cO \left(\frac{\norm{ \xx_0 - \xx^\star}^2}{\gamma T} + \gamma \tilde{L}^2  + \gamma \zeta \tilde{L} \frac{\sens(\mC)\norm{\mB}_F}{\sqrt{T}} \right)
\end{equation}
where $\tilde{L}$ is the Lipshitz constant of $f$. \citet{denisov2022:matrix-fact} therefore use $\sens(\mC)\norm{\mB}_F$ as a proxy for the impact of the factorized noise scheme on convergence. To find factorizations with good convergence properties, \citet{denisov2022:matrix-fact, choquette22:multi-epochs} minimize $\sens(\mC)\norm{\mB}_F$ subject to the constraint $\mA = \mB \mC$, which is equivalent to the following objective:
\begin{problem}[Minimal-Norm Matrix Factorization]\label{prob:frob_objective}
Given a lower triangular matrix $\mA\in\R^{T\times T}$, define $\text{OPT}_F(\mA) = (\mB, \mC)$, where $\mB, \mC \in \R^{T\times T}$ solve the following optimization problem.
\begin{equation}\label{eq:frob_objective}
\min_{\mB, \mC} \norm{\mB}_F^2~~\text{such that}~~\mB\mC = \mA,~\operatorname{sens}(\mC) = 1.
\end{equation}
\end{problem}
\cref{eq:frob_objective} is well-studied in the privacy literature and can be solved with a variety of numerical optimization algorithms~\cite{newton_step_mm,mckenna2021hdmm,denisov2022:matrix-fact,choquette22:multi-epochs}. We also note that \citet{denisov2022:matrix-fact} show that without loss of generality, we can assume $\mB$ and $\mC$ are lower triangular.

\paragraph{Finding improved factorizations.}
We argue that \eqref{eq:prior_rate} is pessimistic in stochastic settings. For SGD (when $\mB = \mA$), the last term in \eqref{eq:prior_rate} is $\cO(\gamma \sens(\mC) \zeta \tilde{L} \sqrt{T})$, which diverges with $T$ for a constant stepsize. However, under the same assumptions as in \citep{denisov2022:matrix-fact}, SGD with constant stepsize actually achieves a faster rate of $\cO(\gamma \sens(\mC)\zeta\tilde{L})$ (see~\citep{ShalevShwartz2009:StochasticCO}). 

In this paper, we turn our attention to the \emph{smooth functions} in order to focus on non-convex functions.
We show in \cref{app:more_examples}, there are matrices $\mB_1, \mB_2$ such that $\sens(\mC_1)\|\mB_1\|_F = \sens(\mC_2)\|\mB_2\|_F$, but \cref{eq:dp-algo} diverges with $\mB_1$ and converges with $\mB_2$, therefore showing that Frobenius norm is not the right measure in the smooth case as well. %

This begs the question of whether there are objectives that better capture the impact of the noise injected in \eqref{eq:dp-algo} on convergence. To answer this, we derive a bound that can exhibit better dependence on $\mB$ to design better factorizations for differentially private optimization.

\section{Problem Formulation}\label{sec:problem_formulation}
To study the effect of the noise $\mB\mZ$ on optimization, we analyze a slightly simplified objective that omits parts of \eqref{eq:dp-algo} not directly related to linear noise correlation. We do this as follows: 
\begin{itemize}
    \item[(I)] We assume that each $\gg_t$ is the true gradient at the point $\xx_t$, i.e. $\gg_t = \nabla f(\xx_t)$.
    \item[(II)] We omit gradient clipping from our analysis. Alternatively, we can view this as setting the clipping threshold $\alpha = \infty$ so that $\overline\mG = \mG$ in \eqref{eq:dp-algo}.
    \item[(III)] We restrict the class $\cA$ to SGD-type algorithms where $\mA = \mS$, as in Example~\ref{ex:prefix-sum}.
\end{itemize}
We impose (I) for simplicity of presentation. Our results can be extended to stochastic gradients in a direct fashion. Restriction (II) is also for simplicity. First, clipping is not directly applied to the noise $\mB\mZ$. Second, for bounded domains or Lipschitz $f$, our analysis still holds with clipping. Last, practical DP methods often use adaptive clipping~\cite{Thakkar19:adaptive_clipping} instead of fixed clipping. We are not aware of convergence analyses for such schemes. We impose (III) in order to limit the class of algorithms $\cA$ to a well-understood subclass. The convergence properties of \eqref{eq:algo} for general matrices $\mA$ are not well-understood even when there is no noise ($\mZ = \0$).
As we discuss in Section~\ref{sec:theory}, even with these simplifications, the effect of $\mB \mZ$ is not well-understood.

Due to (III), we study factorizations $\mB\mC$ of the matrix $\mA = \mS$, as in \cref{ex:prefix-sum}.
Then, \eqref{eq:dp-algo} becomes
\begin{align}\label{eq:opt-setup-matrix}
    \mX = \mX_0 -\gamma \left( \mS \mG + \sens(\mC)\mB \mZ\right).
\end{align}
In vector notation, for $\bb_{0} = \0$ and $\mB = \left[\bb_1, \dots \bb_{T}\right]^\top$,
\begin{align}\label{eq:opt-setup-vector}
    \xx_{t + 1} = \xx_t - \gamma \left[\nabla f(\xx_t) + (\bb_{t + 1} - \bb_{t})^\top\mZ \right],
\end{align}
where for simplicity of presentation, we re-scaled the noise $\mZ$ by the sensitivity, $\sigma^2 = \sens^2(\mC) \zeta^2$. We now discuss several noteworthy special cases of  \eqref{eq:opt-setup-vector}.
\begin{example}[\pgd]\label{ex:pgd}
If $\mB = \mS$ (see \cref{ex:prefix-sum}) we recover SGD with uncorrelated additive noise, also known as \emph{perturbed gradient descent} (\pgd), where
\begin{align}\label{eq:sgd}
     \xx_{t + 1} = \xx_t - \gamma \left[\nabla f(\xx_t) + \zz_{t + 1} \right].
\end{align}
\end{example}
The convergence rate of SGD (and therefore \pgd) is well-understood in the optimization literature (e.g. see \citet[Section 6]{Bubeck15:optimization}). 

\begin{example}[\antipgd]\label{ex:cancel}
By setting $\mB = \mI$, we get an algorithm that at every iteration adds an independent noise vector $\zz_{t + 1}$ and subtracts the previously added noise $\zz_{t}$:
\begin{equation}\label{eq:anti-pgd}
     \xx_{t + 1} = \xx_t - \gamma \left[\nabla f(\xx_t) + \zz_{t + 1} - \zz_{t } \right],~~\zz_{0} = \0
\end{equation}
Intuitively, this removes some of the noise added in the prior round. This is (up to a learning rate factor) the \emph{anti-correlated perturbed gradient descent} (\antipgd) method proposed by \citet{orvieto2022anticorrelated}, who study its generalization properties. Anti-PGD is also equivalent to SGD with randomized-smoothing \cite{duchi2012randomized}. The equivalence follows from defining $\tilde \xx_t = \xx_t + \gamma \zz_t$ and rewriting \eqref{eq:anti-pgd} as
\begin{align*}
     \tilde \xx_{t + 1} = \tilde \xx_t - \gamma \nabla f(\tilde \xx_t - \gamma \zz_t).
\end{align*}
While randomized smoothing algorithm is popular for non-smooth optimization, \citet{vardhan22:smoothing} analyze its convergence properties in the smooth non-convex setting.

\end{example}

\begin{example}[Tree Aggregation \dpftrl]\label{ex:dp-ftrl}
For $k \geq 1$ and $t = 2^{k-1}$, define $\mH_k \in \R^{(2^k-1)\times t}$ recursively as follows:%
\begin{align*}
    \mH_1 = \begin{pmatrix} 1\end{pmatrix},~\mH_{k + 1} = \begin{pmatrix} \mH_k & \0 \\ \0 & \mH_k\\
    \1 & \1\end{pmatrix}
\end{align*}
where $\1$ above represents an all-ones row of appropriate width. For $T = 2^{k-1}$, if $\mC = \mH_k$ and $\mB = \mS\mC_k^{\dagger}$ where $\mC_k^\dagger$ denotes a carefully chosen right pseudo-inverse of $\mC$, then we recover the same noise matrix $\mB$ as in the \dpftrl algorithm with either the online or full Honaker estimator (depending on the choice of $\mC^\dagger$) as in \citep{kairouz21-dp-ftrl,denisov2022:matrix-fact}. 
Note that $\mB, \mC$ are not square. This can be remedied by appropriately projecting onto $\R^T$. See \citet[Appendix D.3]{choquette22:multi-epochs} for details.%
\end{example}

\section{Deriving Tighter Convergence Rates}\label{sec:theory}
We would like convergence rates for \eqref{eq:opt-setup-matrix} that apply to any factorization and yield tight convergence rates for notable special cases. We pay special attention to \pgd (\cref{ex:pgd}) and \antipgd (\cref{ex:cancel}), as they represent extremes in the space of factorizations ($\mS = \mS\mI$ and $\mS = \mI\mS$, respectively). As we will show, it is possible to use existing theoretical tools to derive tight convergence rates for both, \emph{but not simultaneously}.

Below, we discuss ways to derive tight rates for \pgd and \antipgd, and how these rates involve incompatible analyses. We then develop a novel analytic framework involving \emph{restart iterates} that allows us to analyze both methods simultaneously, as well as \eqref{eq:opt-setup-matrix} for general factorizations. We start by formally stating our assumptions. For simplicity of presentation, we re-scale the noise $\mZ$ by the sensitivity of $\mC$, i.e. $\sigma^2 = \sens^2(\mC) \zeta^2$; we will suppress the $\mC$ dependence of $\sigma$.
\begin{assumption}[Noise]\label{as:noise} The rows  $\zz_1, \dots, \zz_T$ of the noise matrix $\mZ$ are independent random vectors such that $\forall t$, $\E[\zz_t] = \0$ and $\E \norm{\zz_t}^2 \leq \sigma^2$.
\end{assumption}
We do not assume $\tilde L$-Lipshitzness in our results, but we do assume $L$-smoothness. This is a relatively standard assumption in optimization literature~\citep{Bubeck15:optimization}.
\begin{assumption}[$L$-smoothness]\label{as:smooth}
The function $f : \R^d \to \R$ is differentiable, and there exists $L > 0$ such that for all $x, y \in \R^d$, $\norm{\nabla f(\xx) - \nabla f(\yy)} \leq L\norm{\xx - \yy}$.
\end{assumption}
For \emph{some} of the results we will assume convexity.
\begin{assumption}[Convexity]\label{as:convex}
The function $f: \R^d \to \R$ is convex, i.e. $\forall \xx, \yy \in \R^d, f(\xx) - f(\yy) \leq \langle \nabla f(\xx), \xx-\yy \rangle$.
When assuming convexity, we also assume the infimum of $f$ is achieved in $\R^d$.
\end{assumption}

\subsection{Convergence Rates for \pgd and \antipgd}
In this section we discuss the (distinct) convergence analyses of \pgd and \antipgd, and the suboptimal results derived by trying to apply the proof technique for one to the other. We focus on the convex setting for brevity, though these analyses can be directly extended to the non-convex setting.

\paragraph{\pgd.} The convergence of \pgd (\cref{ex:pgd}) is well-understood since it is a special case of SGD.
One can show the following.
\begin{proposition}[{Adapted from \citet[Theorem 1]{Dekel12:sgd_convex_proof}}]\label{prop:pgd}
Under Assumptions \ref{as:noise}, \ref{as:smooth} and \ref{as:convex}, if $\mB = \mS$ and $\gamma < \nicefrac{1}{2L}$, then the output of \eqref{eq:opt-setup-matrix} satisfies
\begin{equation}\label{eq:sgd-rate}
    \sum_{t = 0}^T \dfrac{\E \left[f(\xx_t) - f^\star\right]}{T+1} \leq \cO \left( \frac{\norm{ \xx_0 - \xx^\star}^2}{\gamma T} + \gamma \sigma^2  \right).
\end{equation}
\end{proposition}
The proof follows from combining the update \eqref{eq:sgd}, standard facts about convex functions, and the fact that $\gamma < \nicefrac{1}{2L}$, to get the inequality
\begin{align*}
    &\E_t \norm{\xx_{t + 1} - \xx^\star}^2 \leq \norm{\xx_t - \xx^\star}^2 - \gamma \left( f(\xx_t) - f^\star \right) + \gamma^2 \sigma^2.
\end{align*}
It is left to average over iterations $0 \leq t \leq T$.

\paragraph{\antipgd.} For \antipgd (\cref{ex:cancel}), one can show the following.
\begin{proposition}\label{prop:cancel}
Under Assumptions \ref{as:noise}, \ref{as:smooth} and \ref{as:convex}, if $\mB = \mI$ and $\gamma < \nicefrac{1}{2L}$, then the output of \eqref{eq:opt-setup-matrix} satisfies
\begin{align}\label{eq:Anti-PGD-rate}
    \sum_{t = 0}^T \dfrac{\E \left[f(\xx_t) - f^\star\right]}{T+1} \leq \cO \left( \frac{\norm{ \xx_0 - \xx^\star}^2}{\gamma T} + L \gamma^2 \sigma^2 \right)
\end{align}
\end{proposition}
Since $L \gamma < \nicefrac{1}{2}$, the RHS of \eqref{eq:Anti-PGD-rate} is strictly smaller than the RHS of \eqref{eq:sgd-rate}.  While this result may be known, we were unable to find a reference, so we provide a complete proof in \cref{app:cancel}. The proof utilizes perturbed iterate analysis~\citep{mania17:perturbed_analysis}. We define a \emph{virtual sequence} $\{\tilde \xx_t\}_{t = 0}^T$ as follows:
\begin{align}\label{eq:virtual-standard}
    \tilde \xx_{t + 1} = \tilde \xx_t - \gamma \nabla f(\xx_t), && \tilde \xx_0 = \xx_0
\end{align}
The $\tilde\xx_t$ are the iterates of \eqref{eq:opt-setup-matrix} when $\mZ = \0$. We can then prove the following descent inequality:
\begin{align*}
    \norm{\tilde \xx_{t + 1} - \xx^\star}^2 &\leq \norm{\tilde \xx_t - \xx^\star}^2 - \frac{\gamma}{2} \left( f(\xx_t) - f^\star \right) + 2 L \gamma \norm{\tilde \xx_t - \xx_t}^2.
\end{align*}
Because of the anti-correlation in \eqref{eq:anti-pgd}, the virtual iterates $\tilde\xx_t$ are close to the real iterates $\xx_t$, as $\xx_t - \tilde \xx_t = \gamma \zz_t$. Averaging over $t$, 
we recover \eqref{eq:Anti-PGD-rate}. See Appendix~\ref{app:cancel} for details.

\paragraph{Tightness.}
The noise terms (those terms involving $\sigma^2$) in \eqref{eq:sgd-rate}, \eqref{eq:Anti-PGD-rate} are both tight. We show this in \cref{app:noise_lower_bound} on the objective $f(\xx) = (\nicefrac{L}{2})\norm{\xx}^2$. %

\paragraph{Difficulties in a unified analysis.} The proof techniques for \pgd and \antipgd above are notably different, and as we explain in Appendix~\ref{app:difficulty}, do not lead to favorable results when trying to use one of the two strategies to analyze both.

\subsection{Main Results and Analytic Techniques}
To unify the proof techniques above, we use a modified virtual sequence with \emph{restart iterations}. For a parameter $\tau = \tilde\Theta (\nicefrac{1}{L \gamma})$ (throughout, $\tilde{\cO}$ and $\tilde{\Theta}$ hide poly-logarithmic factors), we define
\begin{align}\label{eq:restart}
    \begin{aligned}
        \tilde{\xx}_{t + 1} &= \tilde \xx_t - \gamma \nabla f(\xx_t) \\\tilde \xx_{t + 1} &= \xx_{t + 1}
    \end{aligned}
    &&
    \begin{aligned}
        &\text{if }~t + 1 \neq 0 \bmod \tau \\ &\text{if }~t + 1 = 0 \bmod \tau.
    \end{aligned}
\end{align}
Similar to the virtual sequence in~\eqref{eq:virtual-standard}, $\tilde \xx_t$ incorporates only deterministic gradients $\nabla f(\xx_t)$. However, every $\tau$ iterations we reset $\tilde \xx_t$ to the real iterate $\xx_t$. 
This allows us to control the divergence between the virtual sequence and the real sequence (enabling a tight analysis of \pgd), while still capturing the convergence benefits of anti-correlated noise (enabling a tight analysis of \antipgd).

The parameter $\tau$ is independent of $\mB$, and depends only on the geometry of $f$ and the stepsize $\gamma$. Using this machinery, we can prove convergence rates of \eqref{eq:opt-setup-matrix} for \emph{any} factorization $\mS = \mB\mC$. These rates involve $\ell_2$ distances between the rows $\bb_t$ of the matrix $\mB$ (where $\bb_0 = \mathbf{0}$ for convenience).

\begin{theorem}[non-convex]\label{thm:main_nonconvex}
Suppose Assumptions~\ref{as:noise} and \ref{as:smooth} hold, $\gamma \leq \nicefrac{1}{4L}$, and $\tau = \nicefrac{1}{\gamma L}$. Then \eqref{eq:opt-setup-matrix} produces iterates whose average error $(T+1)^{-1}\sum_{t=0}^T \E\norm{\nabla f(\xx_t)}^2$ is upper bounded by
\begin{align*}
    \cO \Bigg(&\frac{(f(\xx_0) - f^\star)}{\gamma T} + \frac{\sigma^2}{T \tau}\times
    \Bigg[\textstyle \frac{1}{\tau}\sum_{t = 1}^T \norm{\bb_{t} - \bb_{\lfloor \frac{t}{\tau}\rfloor \tau }}^2 + \textstyle\sum_{\substack{1 \leq t \leq T \\ t = 0\bmod{\tau}}} \norm{\bb_{t } - \bb_{t - \tau }}^2\Bigg]\Bigg).
\end{align*}
\end{theorem}

\begin{theorem}[convex]\label{thm:main_convex}
Under Assumptions~\ref{as:noise},~\ref{as:smooth}, and~\ref{as:convex}, if $\gamma \leq \nicefrac{1}{4L}$ and $\tau = \tilde\Theta( \nicefrac{1}{\gamma L})$, then \eqref{eq:opt-setup-matrix} produces iterates with average error $(T+1)^{-1} \sum_{t = 0}^T \E \left[f(\xx_t) - f^\star\right]$ upper bounded by
\begin{align*}
    \tilde\cO\Bigg(&\frac{\norm{ \xx_0 - \xx^\star}^2}{\gamma T} + \frac{\sigma^2 }{T L\tau} \times\Bigg[\textstyle \frac{1}{\tau}\sum_{t = 1}^{T} \norm{\bb_{t} - \bb_{\lfloor \frac{t}{\tau}\rfloor \tau}}^2 + \textstyle \sum_{\substack{1 \leq t \leq T \\ t = 0\bmod{\tau}}} \norm{\bb_{t } - \bb_{t - \tau }}^2 + \norm{\bb_{\lfloor \frac{T}{\tau}\rfloor\tau}}^2\Bigg]\Bigg).
\end{align*}
\end{theorem}
We give complete proofs in Appendix~\ref{app:main_proofs}. These convergence rates consist of two terms: The first term states how fast the function would converge in the absence of the noise. The second term, the \emph{noise term}, is the focus of our paper, as it shows how the correlated noise $\mB \mZ$ affects convergence.

These rates involve only differences of rows of $\mB$ that are at most $\tau$ iterations apart. Intuitively, $\tau$ is a coarse indicator of whether an iterate $\xx_t$ is still sensitive to the noise injected at an iteration $t' < t$. If $t > t' + \tau$, then changes in the noise added at step $t$ are effectively uncorrelated to iteration $t'$. As we detail in Appendix, applying Theorem \ref{thm:main_convex} to the special cases in Examples~\ref{ex:pgd}, \ref{ex:cancel} recovers their tight convergence rates in \eqref{eq:sgd-rate}, \eqref{eq:Anti-PGD-rate} correspondingly.

\section{Finding Better Factorizations}\label{sec:better_factorization}
We now draw on our results in \cref{sec:theory} to develop better mechanisms for the \dpmf framework. We modify the objective underlying the offline matrix factorization problem during the first stage of the \dpmf workflow (\cref{fig:mf_workflow}).  
Specifically, observe that
the noise term in Theorems~\ref{thm:main_nonconvex} and \ref{thm:main_convex} can be rewritten in matrix notation (up to multiplicative constants) as
\begin{equation} \label{eq:modfiedobj}
\resizebox{0.9\hsize}{!}{
    $\norm{\mLambda_{\tau} \mB}_F^2 = \sum_{t = 1}^T \norm{\mlambda_t^\top \mB}^2 =  \textstyle{\sum_{\substack{1 \leq t \leq T \\ t = 0\bmod{\tau}}}}\norm{\bb_{t} - \bb_{t - \tau}}^2 + {\textstyle\sum_{\substack{1 \leq t \leq T \\ t \neq 0\bmod{\tau} }}} \norm{\frac{1}{\sqrt{\tau}}\left(\bb_{t} - \bb_{\lfloor \frac{t}{\tau}\rfloor \tau}\right)}^2$}
\end{equation}
where $\mLambda_{\tau} = \left[\mlambda_1^\top, \dots, \mlambda_T^\top \right]^\top \in \R^{T \times T}$, and we set the rows $\mlambda_t$ appropriately to select corresponding row differences of $\mB$ with either coefficient $1$ or $\nicefrac{1}{\sqrt{\tau}}$ depending on the index $t$. We give a precise definition of $\mLambda_\tau$ and an explicit example when $T = 12, \tau = 3$ in Appendix~\ref{app:weight_matrix}.%

Recall that \cite{denisov2022:matrix-fact} minimize the Frobenius norm objective \eqref{eq:frob_objective} based on their derived convergence bounds in \eqref{eq:prior_rate}.
Since our derived convergence bounds are strictly tighter, we propose using \cref{eq:modfiedobj} as the new objective function in \eqref{eq:frob_objective}. Intuitively, since $\|\mLambda_{\tau} \mB\|_F^2$ is a better proxy for learning performance than $\|\mB\|_F^2$, minimizing this quantity in the offline factorization problem should lead to ERM methods with better privacy-utility trade-offs.

We can solve our new offline matrix factorization problem in a straightforward manner. We can show that for $\mA=\mS$, we can solve this modified problem by first computing the solution $\tilde\mB, \tilde\mC$ using $\text{OPT}_F(\mLambda_\tau\mA)$. The solution to our modified objective is then $\mC = \tilde\mC$, $\mB = \mA\mC^{-1}$. This implies we can use existing open-source solvers designed for \eqref{eq:frob_objective}~\cite{newton_step_mm,mckenna2021hdmm,denisov2022:matrix-fact}.

\section{Experiments}\label{sec:exp}

In this section, we evaluate the ERM performance of \dpmf under different offline factorization objectives. We focus on the Frobenius norm objective \eqref{eq:frob_objective}, which we refer to as DP-MF  \cite{denisov2022:matrix-fact, choquette22:multi-epochs}, and our modified objective \eqref{eq:modfiedobj}, which we refer to as DP-MF$^+$.

\subsection{Validating Theoretical Results}

\begin{figure*}[tb]
\centering     %
\subfigure[Average gradient norm for varying learning rates.]{\label{fig:avg}\includegraphics[width=40mm]{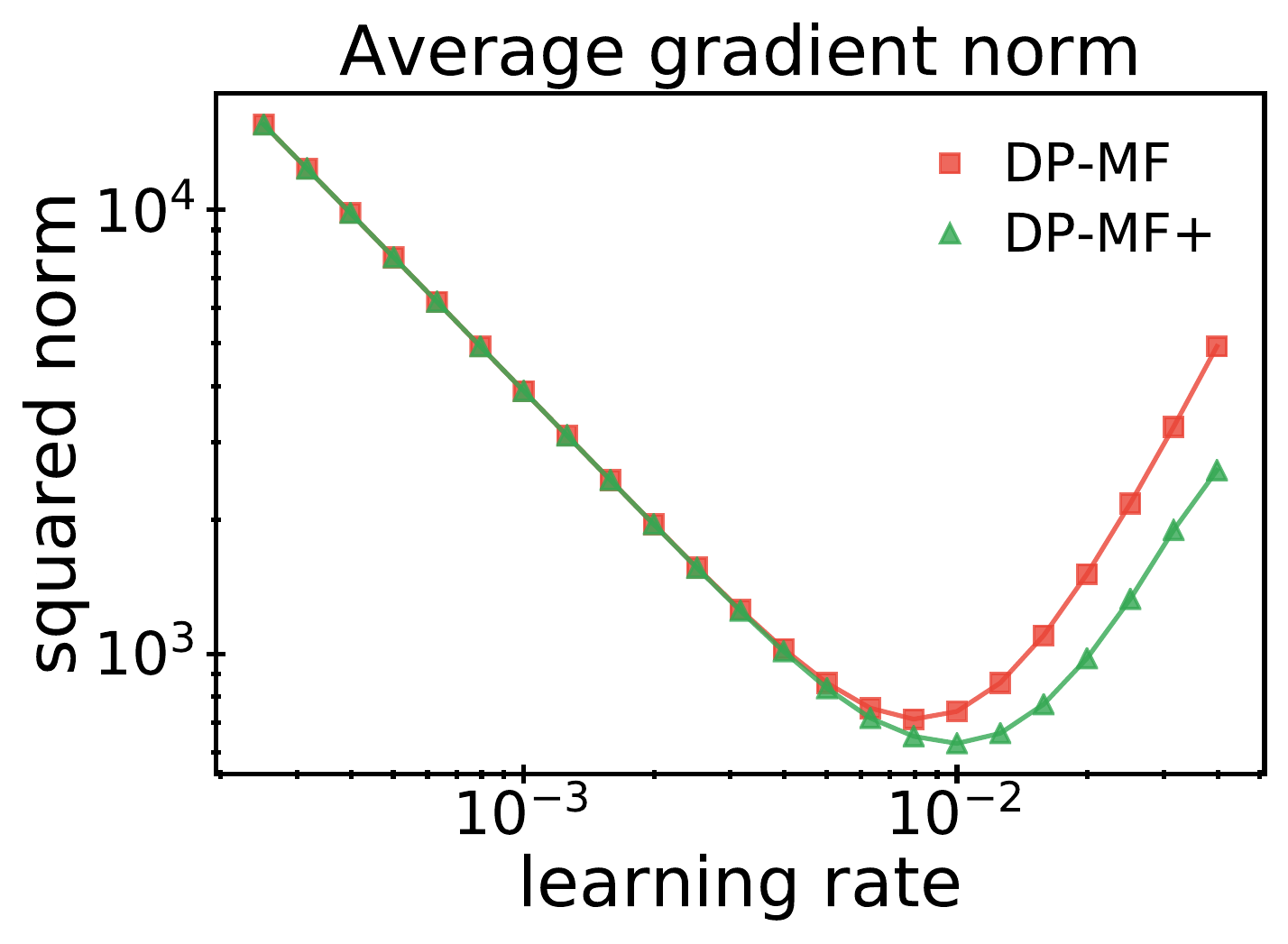}}
\subfigure[Last gradient norm for varying learning rates.]{\label{fig:final}\includegraphics[width=40mm]{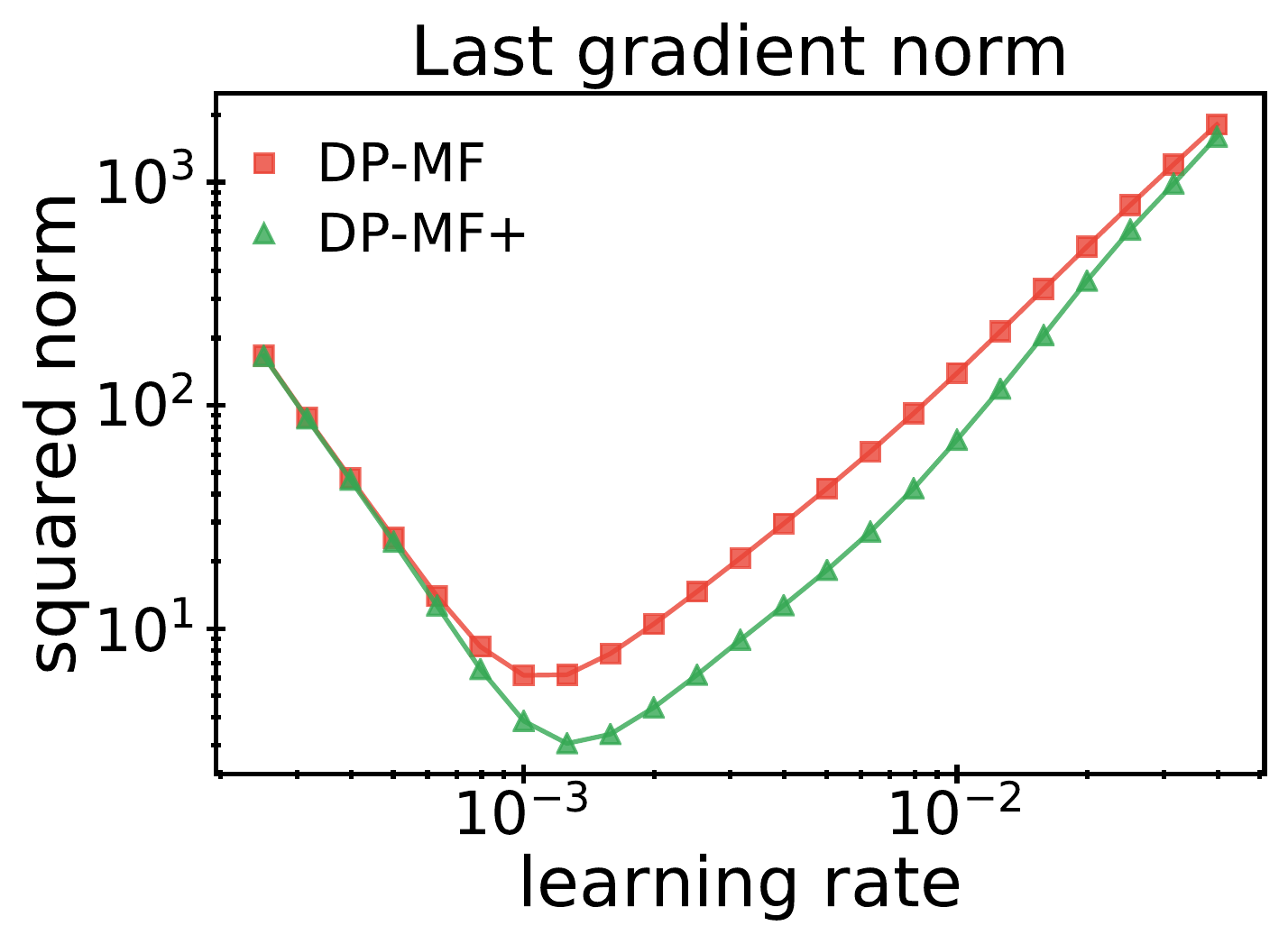}}
\subfigure[Gradient norm over time for $\gamma = 10^{-2}$.]{\label{fig:convergence}\includegraphics[width=40mm]{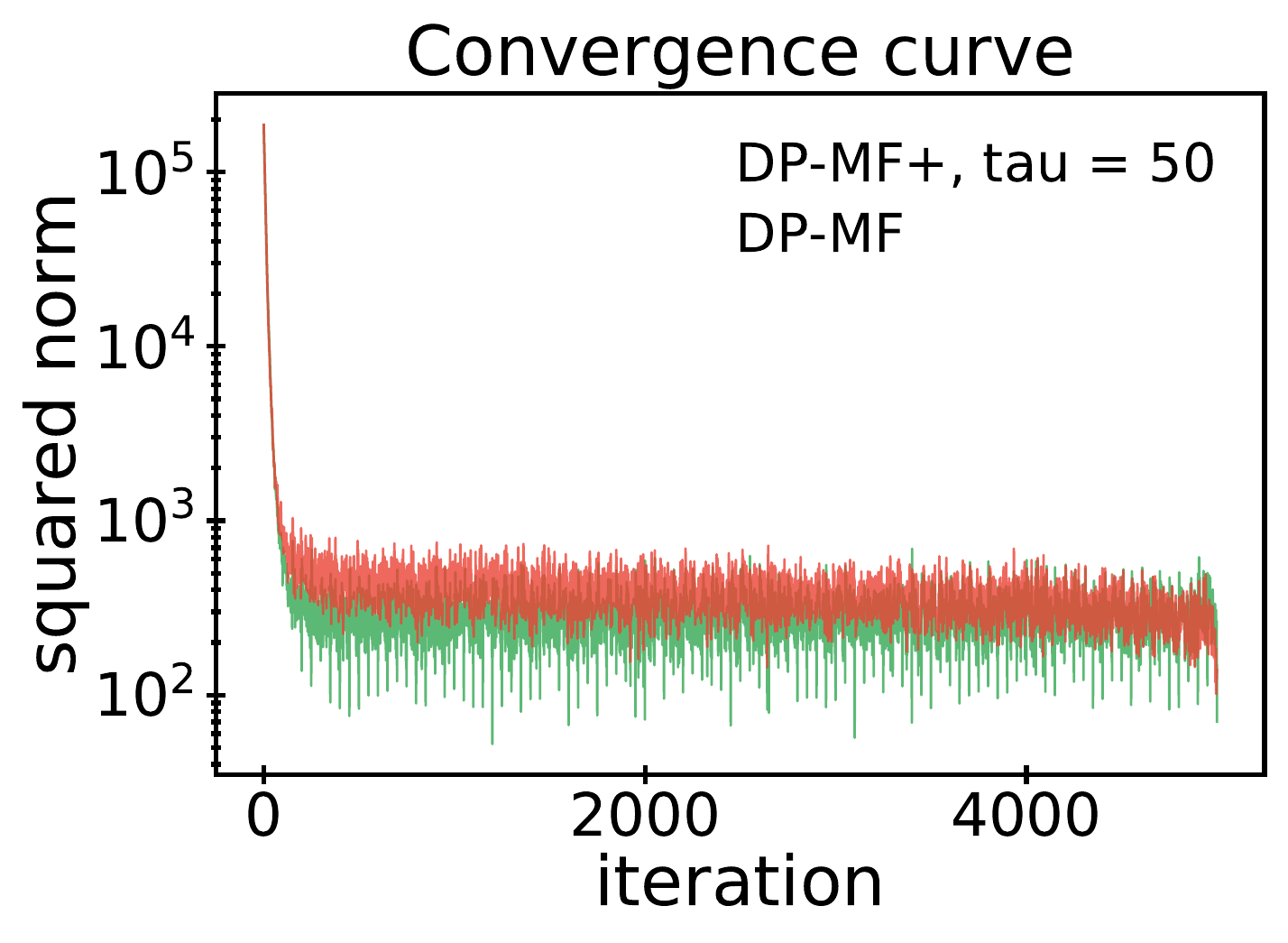}}
\caption{Comparison of the average and last gradient norms for DP-MF and DP-MF$^+$ on a random non-strongly convex quadratic function with $L = 10$.}\label{fig:avg_gradient_norm_main}
\end{figure*}

We first validate our theoretical results above by comparing the convergence of DP-MF and DP-MF$^+$ on a \emph{random quadratic} function that satisfies the assumptions of \cref{thm:main_convex}. Notably, we ensure the quadratic is not strongly convex. We treat $\tau$ in \eqref{eq:modfiedobj} as a hyperparameter and tune it over a fixed grid. For complete details, please refer to Appendix~\ref{app:more_experiments}. We present the results in \cref{fig:avg_gradient_norm_main}.

In \cref{fig:avg} we plot $\frac{1}{T} \sum_{t = 0}^T \norm{\nabla f(\xx_t)}^2$, as this quantity is proportional to the LHS of Theorem~\ref{thm:main_convex}. %
For all learning rates, DP-MF$^+$ either matches or outperforms DP-MF. Moreover, the advantage of DP-MF$^+$ increases as the learning rate increases. This corresponds to our theory in \cref{thm:main_convex}. Indeed, the larger the stepsize $\gamma$, the smaller the optimal $\tau$ (as $\tau = \Theta\left(\nicefrac{1}{\gamma L}\right)$), and the more often restarts are used in the analysis of Theorem~\ref{thm:main_convex}.

\cref{fig:final} further depicts the last-iterate behaviours of DP-MF and DP-MF$^+$, which is often more practically relevant. Interestingly, the last iterate behaviour is improved even in the cases where the average behaviour does not improve.
Finally, in \cref{fig:convergence} we pick $\gamma = 10^{-2}, \tau = 50$ as the parameters for which both the average and the last-iterate behaviours are improved and plot the convergence curve over iterations. DP-MF$^+$ has regular oscillating behaviour, allowing it to achieve a good final-iterate performance. The period of these oscillations is exactly equal to $\tau$.

\subsection{Practical DP Training Experiments}

We now compare DP-MF, DP-MF$^+$, and DP-SGD with privacy amplification~\citep{dpsgd_2016} on the MNIST, CIFAR-10, and Stack Overflow datasets. We omit from comparison DP-FTRL \cite{kairouz21-dp-ftrl} and DP-Fourier \cite{choquette22:multi-epochs} as these methods are strictly dominated by DP-MF. Unlike our theoretical analysis, we include clipping to derive formal $(\epsilon, \delta)$ privacy guarantees. To facilitate a fair comparison, we set $\delta = 10^{-6}$ in all the settings, and compare against varying $\epsilon$.
We give complete experimental details in Appendix~\ref{app:more_experiments}

\begin{figure*}[tb]
\centering     %
\subfigure[MNIST, $1$ epoch]{\label{fig:mnist1}\includegraphics[width=50mm]{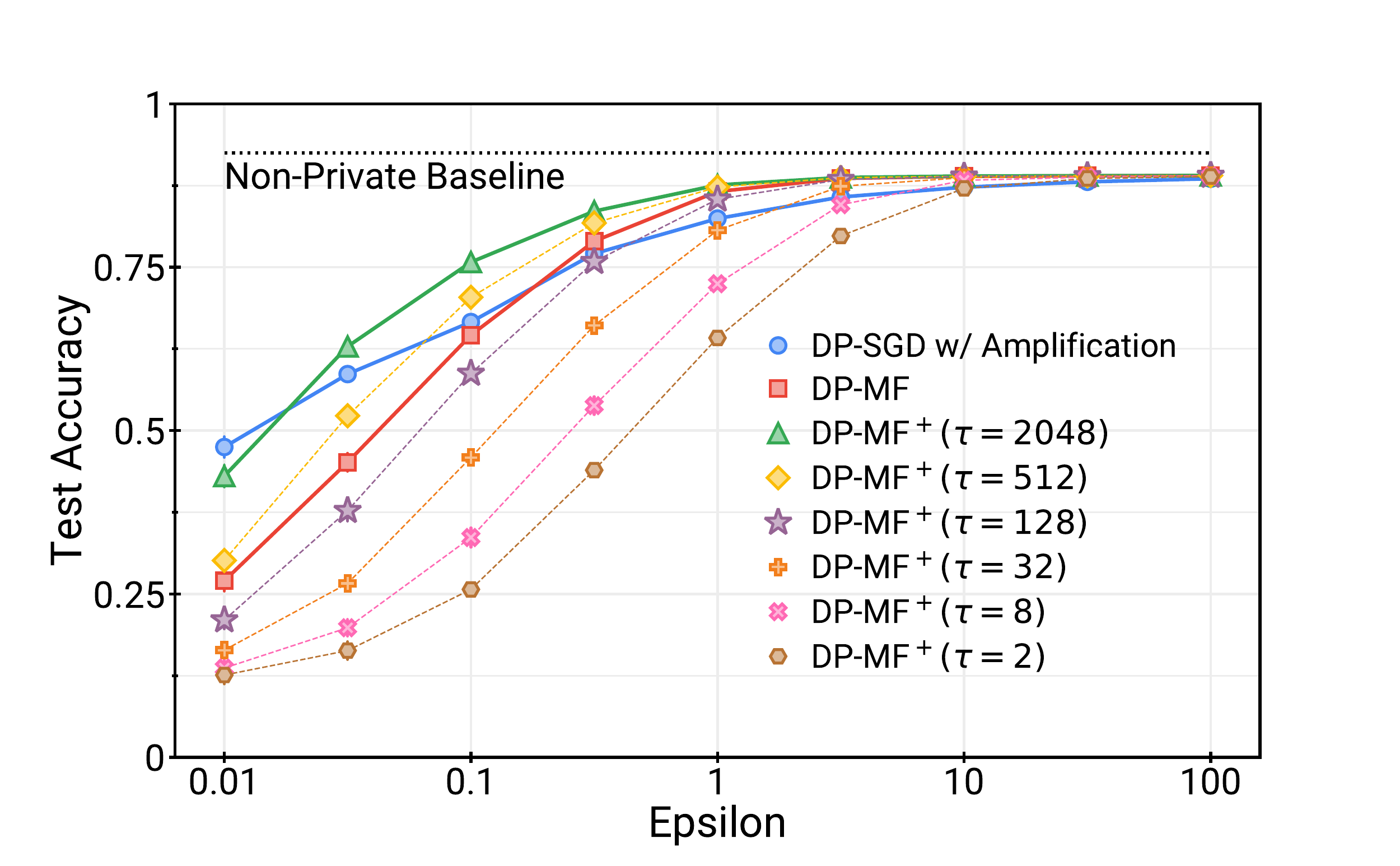}}\hspace{-2em}
\subfigure[MNIST, $16$ epochs]{\label{fig:mnist16}\includegraphics[width=50mm]{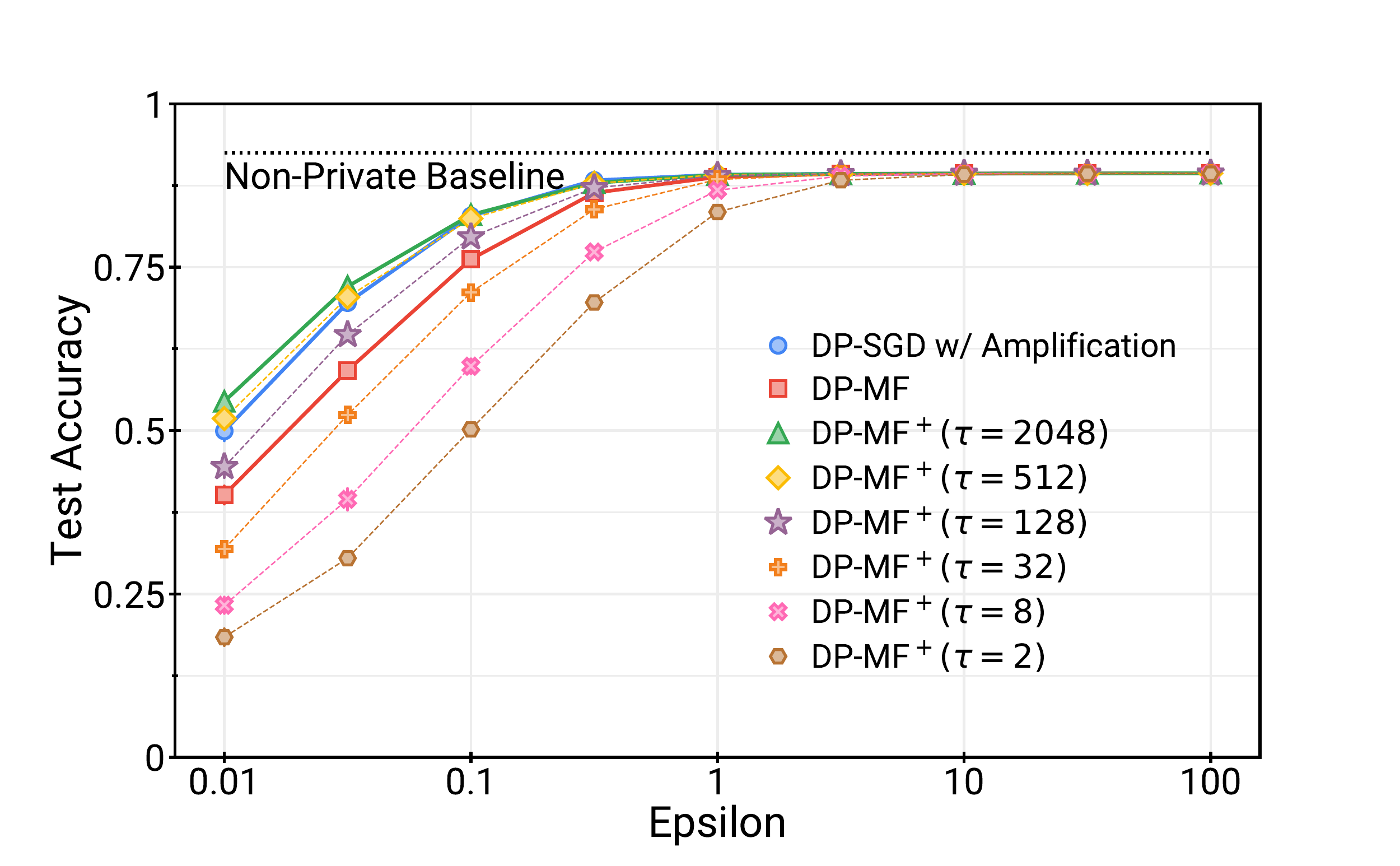}}\hspace{-2em}
\subfigure[CIFAR-10, $20$ epochs]{\label{fig:cifar20}\includegraphics[width=50mm]{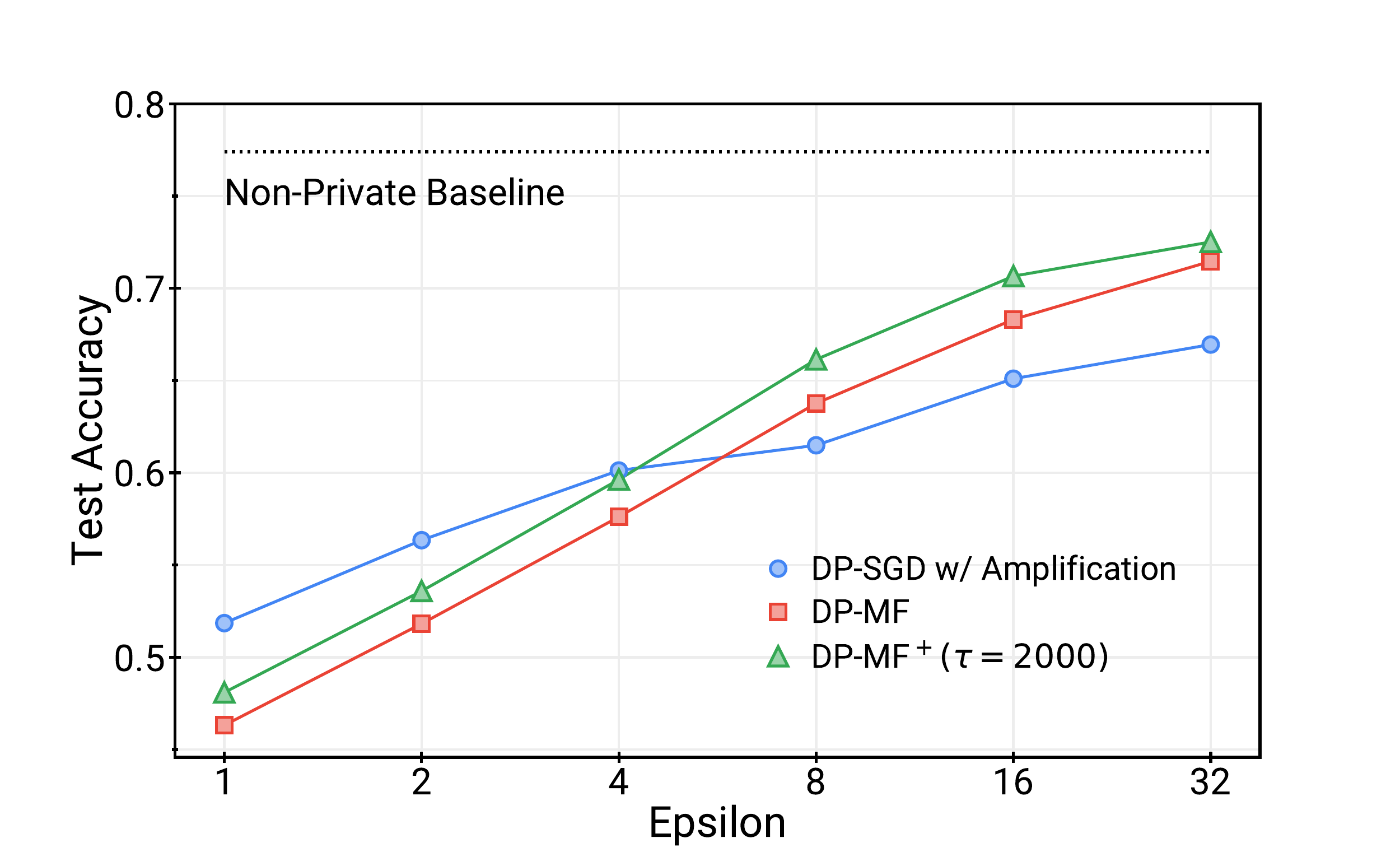}}
\caption{Test set accuracy of various mechanisms on the MNIST and CIFAR-10 datasets.}
\end{figure*}

\paragraph*{MNIST, logistic regression.} We train for $T = 2048$ iterations and either $1$ or $16$ epochs depending on the batch size, corresponding to a batch size of $29$ and $469$ respectively.\footnote{In practice, one often trains small-scale models for many epochs, perhaps even using full-batch gradients, to improve the privacy/utility trade-off (at the cost of increased computation). We are interested in the \emph{relative} performance for a fixed computation budget, so we train for a small number of epochs.} We fix the clipping threshold at $1.0$ and the learning rate at $0.5$.
We vary $\tau$ in \eqref{eq:modfiedobj} over $\{2, 8, 32, 128, 512, 2048\}$. The results are in \cref{fig:mnist1,fig:mnist16}. DP-MF$^+$ improves monotonically with $\tau$, performing best when $\tau = 2048 = T$. For such $\tau$, DP-MF$^+$ consistently out-performs DP-MF across all settings. Recall from \eqref{eq:modfiedobj} that this corresponds to the offline objective $\|\mLambda_T \mB\|_F^2 $ where $\lambda_{ii} = \nicefrac{1}{\sqrt{T}}$ for all $i < T$ and $\lambda_{T T} = 1$. This objective strongly penalizes errors on the final iterate, which is the model used to compute test accuracy.

We also see that DP-MF$^+$ expands the number of settings in which we can beat DP-SGD. DP-MF only outperforms DP-SGD for sufficiently large $\epsilon$ ($\epsilon \geq 0.31$ for $1$ epoch and $\epsilon \geq 31$ for $16$ epochs). By contrast, DP-MF$^+$ outperforms DP-SGD in every setting except when $\epsilon = 0.01$ and $1$ epoch. None of the mechanisms reached the accuracy levels obtained by the non-private baseline, even at $\epsilon = 100$.  We suspect this is due to the fact that we are using a fixed but aggressive clipping threshold of $1.0$ across all experiments, which helps in the moderate privacy regime but hurts in very low privacy regime. Even though DP-MF$^+$ does not use privacy amplification, it outperforms  DP-SGD, which uses privacy amplification. This is due to the efficient noise anti-correlations in DP-MF$^+$. If amplification were not possible, performance of DP-SGD would degrade even further.

\paragraph*{CIFAR-10, CNN.} We follow the experimental setup from \citep{choquette22:multi-epochs}.  Specifically, we train all mechanisms for $20$ epochs and $T = 2000$ iterations, which corresponds to a batch size of $500$.\footnote{While \citet{choquette22:multi-epochs} use momentum and learning rate decay, we omit the use of such techniques as they are orthogonal to our theoretical results.} We tune the learning rate over a fixed grid. 
We fix $\tau = T = 2000$ in DP-MF$^+$ as we found that worked best in the MNIST experiments. The results are given in \cref{fig:cifar20}. We see that DP-MF$^+(\tau=2000)$ offers a consistent improvement over DP-MF across all choices of $\epsilon$ considered. Both DP-MF and DP-MF$^+$ beat DP-SGD for $\epsilon>4$.  This observation is consistent with prior work on \dpftrl and DP-MF, where DP-SGD performs relatively better with smaller $\epsilon$ while DP-MF performs better with larger $\epsilon$.

\paragraph*{Stack Overflow, LSTM.} In Appendix \ref{app:more_experiments}, we compare DP-MF and DP-MF$^+$ on a federated learning task with \emph{user-level} differential privacy. We do not compare DP-SGD on this task, as amplification techniques such as shuffling and subsampling are not possible in practical federated learning settings~\cite{kairouz21-dp-ftrl}. In this task, we train an LSTM network to do next-word prediction on the Stack Overflow dataset. To be consistent with the prior work \cite{denisov2022:matrix-fact} and to test if our proposed factorizations are compatible with the other types of workloads $\mA$ from \cref{eq:algo}, we use momentum and learning rate decay. Our results are given in \cref{tab:stackoverflow}.  We see that two methods perform comparably, verifying competitiveness of our method. Note that this task uses federated averaging~\cite{mcmahan2017communication} instead of gradient descent. Developing offline factorization objectives specifically for federated learning remains an open problem.

\section{Conclusion}
In this work, we developed analytic techniques to study the convergence of gradient descent under linearly correlated noise that is motivated from a class of DP mechanisms. We derived tighter bounds than currently exist in the literature, and we use our novel theoretical understanding to design privacy mechanisms with improved convergence. Perhaps more importantly, our work highlights the wealth of stochastic optimization questions arising from recent advances in differentially private model training.  As such, we distill and formalize various optimization problems arising from recent work on matrix mechanisms for DP. %
Our work raises a host of questions and open problems, including extending our analysis to include things such as clipping, shuffling, and momentum. Another key extension is to derive last-iterate convergence rates rather than average-iterate convergence rates, as in some settings it is only the final ``released'' model that needs formal privacy guarantees. Given the improved generalization properties of \antipgd~\citep{orvieto2022anticorrelated}, one could also investigate how to design more general linearly correlated noise mechanisms which improve both privacy and generalization.

\section{Acknowledgments}
The authors would like to thank Francesco D'Angelo, Nina Mainusch and Linara Adylova for their comments on the manuscript. The authors would also like to thank the reviewers for their helpful suggestions in improving the clarity of the writing.

\bibliographystyle{plainnat}
\bibliography{references}

\begin{thebibliography}{55}
\providecommand{\natexlab}[1]{#1}
\providecommand{\url}[1]{\texttt{#1}}
\expandafter\ifx\csname urlstyle\endcsname\relax
  \providecommand{\doi}[1]{doi: #1}\else
  \providecommand{\doi}{doi: \begingroup \urlstyle{rm}\Url}\fi

\bibitem[Abadi et~al.(2016)Abadi, Chu, Goodfellow, McMahan, Mironov, Talwar,
  and Zhang]{dpsgd_2016}
Martin Abadi, Andy Chu, Ian Goodfellow, H.~Brendan McMahan, Ilya Mironov, Kunal
  Talwar, and Li~Zhang.
\newblock Deep learning with differential privacy.
\newblock \emph{Proceedings of the 2016 ACM SIGSAC Conference on Computer and
  Communications Security}, Oct 2016.
\newblock \doi{10.1145/2976749.2978318}.
\newblock URL \url{http://dx.doi.org/10.1145/2976749.2978318}.

\bibitem[Agarwal and Duchi(2011)]{Agarwal11:delayedSGD}
Alekh Agarwal and John~C Duchi.
\newblock Distributed delayed stochastic optimization.
\newblock In J.~Shawe-Taylor, R.~Zemel, P.~Bartlett, F.~Pereira, and K.Q.
  Weinberger, editors, \emph{Advances in Neural Information Processing
  Systems}, volume~24. Curran Associates, Inc., 2011.
\newblock URL
  \url{https://proceedings.neurips.cc/paper/2011/file/f0e52b27a7a5d6a1a87373dffa53dbe5-Paper.pdf}.

\bibitem[Authors(2019)]{stackoverflow}
The TensorFlow~Federated Authors.
\newblock Tensor{F}low {F}ederated {Stack Overflow} dataset, 2019.
\newblock URL
  \url{https://www.tensorflow.org/federated/api_docs/python/tff/simulation/datasets/stackoverflow/load_data}.

\bibitem[Bassily et~al.(2014)Bassily, Smith, and Thakurta]{bassily2014private}
Raef Bassily, Adam Smith, and Abhradeep Thakurta.
\newblock Private empirical risk minimization: Efficient algorithms and tight
  error bounds.
\newblock In \emph{2014 IEEE 55th annual symposium on foundations of computer
  science}, pages 464--473. IEEE, 2014.

\bibitem[Bassily et~al.(2019)Bassily, Feldman, Talwar, and
  Guha~Thakurta]{bassily2019private}
Raef Bassily, Vitaly Feldman, Kunal Talwar, and Abhradeep Guha~Thakurta.
\newblock Private stochastic convex optimization with optimal rates.
\newblock \emph{Advances in neural information processing systems}, 32, 2019.

\bibitem[Bubeck(2015)]{Bubeck15:optimization}
S\'{e}bastien Bubeck.
\newblock Convex optimization: Algorithms and complexity.
\newblock \emph{Found. Trends Mach. Learn.}, 8\penalty0 (3–4):\penalty0
  231–357, nov 2015.
\newblock ISSN 1935-8237.
\newblock \doi{10.1561/2200000050}.
\newblock URL \url{https://doi.org/10.1561/2200000050}.

\bibitem[Choquette-Choo et~al.(2022)Choquette-Choo, McMahan, Rush, and
  Thakurta]{choquette22:multi-epochs}
Christopher~A. Choquette-Choo, H.~Brendan McMahan, Keith Rush, and Abhradeep
  Thakurta.
\newblock Multi-epoch matrix factorization mechanisms for private machine
  learning, 2022.
\newblock URL \url{https://arxiv.org/abs/2211.06530}.

\bibitem[Das et~al.(2022)Das, Kale, Xu, Zhang, and Sanghavi]{das2022beyond}
Rudrajit Das, Satyen Kale, Zheng Xu, Tong Zhang, and Sujay Sanghavi.
\newblock Beyond uniform {L}ipschitz condition in differentially private
  optimization.
\newblock \emph{arXiv preprint arXiv:2206.10713}, 2022.

\bibitem[Dekel et~al.(2012)Dekel, Gilad-Bachrach, Shamir, and
  Xiao]{Dekel12:sgd_convex_proof}
Ofer Dekel, Ran Gilad-Bachrach, Ohad Shamir, and Lin Xiao.
\newblock Optimal distributed online prediction using mini-batches.
\newblock \emph{J. Mach. Learn. Res.}, 13\penalty0 (null):\penalty0 165–202,
  jan 2012.
\newblock ISSN 1532-4435.

\bibitem[Denisov et~al.(2022)Denisov, McMahan, Rush, Smith, and
  Thakurta]{denisov2022:matrix-fact}
Sergey Denisov, Brendan McMahan, Keith Rush, Adam Smith, and Abhradeep~Guha
  Thakurta.
\newblock Improved differential privacy for {SGD} via optimal private linear
  operators on adaptive streams.
\newblock In \emph{Neural Information Processing Systems}, 2022.

\bibitem[Duchi et~al.(2012)Duchi, Bartlett, and
  Wainwright]{duchi2012randomized}
John~C Duchi, Peter~L Bartlett, and Martin~J Wainwright.
\newblock Randomized smoothing for stochastic optimization.
\newblock \emph{SIAM Journal on Optimization}, 22\penalty0 (2):\penalty0
  674--701, 2012.

\bibitem[Dutta et~al.(2018)Dutta, Joshi, Ghosh, Dube, and
  Nagpurkar]{dutta2018slow}
Sanghamitra Dutta, Gauri Joshi, Soumyadip Ghosh, Parijat Dube, and Priya
  Nagpurkar.
\newblock Slow and stale gradients can win the race: Error-runtime trade-offs
  in distributed sgd.
\newblock In \emph{International conference on artificial intelligence and
  statistics}, pages 803--812. PMLR, 2018.

\bibitem[Dwork et~al.(2006)Dwork, McSherry, Nissim, and Smith]{dp_def}
Cynthia Dwork, Frank McSherry, Kobbi Nissim, and Adam Smith.
\newblock Calibrating noise to sensitivity in private data analysis.
\newblock In Shai Halevi and Tal Rabin, editors, \emph{Theory of Cryptography},
  pages 265--284, Berlin, Heidelberg, 2006. Springer Berlin Heidelberg.
\newblock ISBN 978-3-540-32732-5.

\bibitem[Edmonds et~al.(2020)Edmonds, Nikolov, and Ullman]{edmonds2020power}
Alexander Edmonds, Aleksandar Nikolov, and Jonathan Ullman.
\newblock The power of factorization mechanisms in local and central
  differential privacy.
\newblock In \emph{Proceedings of the 52nd Annual ACM SIGACT Symposium on
  Theory of Computing}, pages 425--438, 2020.

\bibitem[Erlingsson et~al.(2019)Erlingsson, Feldman, Mironov, Raghunathan,
  Talwar, and Thakurta]{erlingsson2019amplification}
{\'U}lfar Erlingsson, Vitaly Feldman, Ilya Mironov, Ananth Raghunathan, Kunal
  Talwar, and Abhradeep Thakurta.
\newblock Amplification by shuffling: From local to central differential
  privacy via anonymity.
\newblock In \emph{Proceedings of the Thirtieth Annual ACM-SIAM Symposium on
  Discrete Algorithms}, pages 2468--2479. SIAM, 2019.

\bibitem[Feldman et~al.(2022)Feldman, McMillan, and Talwar]{feldman2022hiding}
Vitaly Feldman, Audra McMillan, and Kunal Talwar.
\newblock Hiding among the clones: A simple and nearly optimal analysis of
  privacy amplification by shuffling.
\newblock In \emph{2021 IEEE 62nd Annual Symposium on Foundations of Computer
  Science (FOCS)}, pages 954--964. IEEE, 2022.

\bibitem[Gorbunov et~al.(2020)Gorbunov, Kovalev, Makarenko, and
  Richt{\'a}rik]{gorbunov2020linearly}
Eduard Gorbunov, Dmitry Kovalev, Dmitry Makarenko, and Peter Richt{\'a}rik.
\newblock Linearly converging error compensated sgd.
\newblock \emph{Advances in Neural Information Processing Systems},
  33:\penalty0 20889--20900, 2020.

\bibitem[Henzinger and Upadhyay(2022)]{constant_matters}
Monika Henzinger and Jalaj Upadhyay.
\newblock Constant matters: Fine-grained complexity of differentially private
  continual observation using completely bounded norms.
\newblock Cryptology ePrint Archive, Paper 2022/225, 2022.
\newblock URL \url{https://eprint.iacr.org/2022/225}.
\newblock \url{https://eprint.iacr.org/2022/225}.

\bibitem[Jin et~al.(2021)Jin, Netrapalli, Ge, Kakade, and
  Jordan]{jin2021nonconvex}
Chi Jin, Praneeth Netrapalli, Rong Ge, Sham~M Kakade, and Michael~I Jordan.
\newblock On nonconvex optimization for machine learning: Gradients,
  stochasticity, and saddle points.
\newblock \emph{Journal of the ACM (JACM)}, 68\penalty0 (2):\penalty0 1--29,
  2021.

\bibitem[Kairouz et~al.(2021{\natexlab{a}})Kairouz, Mcmahan, Song, Thakkar,
  Thakurta, and Xu]{kairouz21-dp-ftrl}
Peter Kairouz, Brendan Mcmahan, Shuang Song, Om~Thakkar, Abhradeep Thakurta,
  and Zheng Xu.
\newblock Practical and private (deep) learning without sampling or shuffling.
\newblock In Marina Meila and Tong Zhang, editors, \emph{Proceedings of the
  38th International Conference on Machine Learning}, volume 139 of
  \emph{Proceedings of Machine Learning Research}, pages 5213--5225. PMLR,
  18--24 Jul 2021{\natexlab{a}}.
\newblock URL \url{https://proceedings.mlr.press/v139/kairouz21b.html}.

\bibitem[Kairouz et~al.(2021{\natexlab{b}})Kairouz, McMahan, Avent, Bellet,
  Bennis, Bhagoji, Bonawitz, Charles, Cormode, Cummings,
  et~al.]{kairouz2021advances}
Peter Kairouz, H~Brendan McMahan, Brendan Avent, Aur{\'e}lien Bellet, Mehdi
  Bennis, Arjun~Nitin Bhagoji, Kallista Bonawitz, Zachary Charles, Graham
  Cormode, Rachel Cummings, et~al.
\newblock Advances and open problems in federated learning.
\newblock \emph{Foundations and Trends{\textregistered} in Machine Learning},
  14\penalty0 (1--2):\penalty0 1--210, 2021{\natexlab{b}}.

\bibitem[Karimireddy et~al.(2020)Karimireddy, Kale, Mohri, Reddi, Stich, and
  Suresh]{karimireddy2020scaffold}
Sai~Praneeth Karimireddy, Satyen Kale, Mehryar Mohri, Sashank Reddi, Sebastian
  Stich, and Ananda~Theertha Suresh.
\newblock Scaffold: Stochastic controlled averaging for federated learning.
\newblock In \emph{International Conference on Machine Learning}, pages
  5132--5143. PMLR, 2020.

\bibitem[Koloskova et~al.(2020)Koloskova, Loizou, Boreiri, Jaggi, and
  Stich]{koloskova20:unified}
Anastasia Koloskova, Nicolas Loizou, Sadra Boreiri, Martin Jaggi, and
  Sebastian~U. Stich.
\newblock A unified theory of decentralized {SGD} with changing topology and
  local updates.
\newblock In \emph{Proceedings of the 37th International Conference on Machine
  Learning}, ICML'20. JMLR.org, 2020.

\bibitem[Koskela et~al.(2021)Koskela, J{\"a}lk{\"o}, Prediger, and
  Honkela]{koskela2021tight}
Antti Koskela, Joonas J{\"a}lk{\"o}, Lukas Prediger, and Antti Honkela.
\newblock Tight differential privacy for discrete-valued mechanisms and for the
  subsampled gaussian mechanism using {FFT}.
\newblock In \emph{International Conference on Artificial Intelligence and
  Statistics}, pages 3358--3366. PMLR, 2021.

\bibitem[Li et~al.(2010)Li, Hay, Rastogi, Miklau, and
  McGregor]{li2010optimizing}
Chao Li, Michael Hay, Vibhor Rastogi, Gerome Miklau, and Andrew McGregor.
\newblock Optimizing linear counting queries under differential privacy.
\newblock In \emph{Proceedings of the twenty-ninth ACM SIGMOD-SIGACT-SIGART
  symposium on Principles of database systems}, pages 123--134, 2010.

\bibitem[Li et~al.(2015)Li, Miklau, Hay, Mcgregor, and Rastogi]{Li2015TheMM}
Chao Li, Gerome Miklau, Michael Hay, Andrew Mcgregor, and Vibhor Rastogi.
\newblock The matrix mechanism: optimizing linear counting queries under
  differential privacy.
\newblock \emph{The VLDB Journal}, 24:\penalty0 757--781, 2015.

\bibitem[Lucchi et~al.(2022)Lucchi, Proske, Orvieto, Bach, and
  Kersting]{lucchi22:noise_correlation}
Aurelien Lucchi, Frank Proske, Antonio Orvieto, Francis Bach, and Hans
  Kersting.
\newblock On the theoretical properties of noise correlation in stochastic
  optimization.
\newblock \emph{Neural Information Processing Systems}, 2022.

\bibitem[Mania et~al.(2017)Mania, Pan, Papailiopoulos, Recht, Ramchandran, and
  Jordan]{mania17:perturbed_analysis}
Horia Mania, Xinghao Pan, Dimitris Papailiopoulos, Benjamin Recht, Kannan
  Ramchandran, and Michael~I. Jordan.
\newblock Perturbed iterate analysis for asynchronous stochastic optimization.
\newblock \emph{SIAM Journal on Optimization}, 27\penalty0 (4):\penalty0
  2202--2229, 2017.
\newblock \doi{10.1137/16M1057000}.
\newblock URL \url{https://doi.org/10.1137/16M1057000}.

\bibitem[McKenna et~al.(2018)McKenna, Miklau, Hay, and
  Machanavajjhala]{mckenna2018optimizing}
Ryan McKenna, Gerome Miklau, Michael Hay, and Ashwin Machanavajjhala.
\newblock Optimizing error of high-dimensional statistical queries under
  differential privacy.
\newblock \emph{arXiv preprint arXiv:1808.03537}, 2018.

\bibitem[McKenna et~al.(2021)McKenna, Miklau, Hay, and
  Machanavajjhala]{mckenna2021hdmm}
Ryan McKenna, Gerome Miklau, Michael Hay, and Ashwin Machanavajjhala.
\newblock Hdmm: Optimizing error of high-dimensional statistical queries under
  differential privacy.
\newblock \emph{arXiv preprint arXiv:2106.12118}, 2021.

\bibitem[McMahan and Thakurta(2022)]{mcmahan2022federated}
Brendan McMahan and Abhradeep Thakurta.
\newblock Federated learning with formal differential privacy guarantees.
\newblock \emph{Google AI Blog}, 2022.

\bibitem[McMahan et~al.(2017)McMahan, Moore, Ramage, Hampson, and
  y~Arcas]{mcmahan2017communication}
Brendan McMahan, Eider Moore, Daniel Ramage, Seth Hampson, and Blaise~Aguera
  y~Arcas.
\newblock Communication-efficient learning of deep networks from decentralized
  data.
\newblock In \emph{Artificial intelligence and statistics}, pages 1273--1282.
  PMLR, 2017.

\bibitem[Mishchenko et~al.(2020)Mishchenko, Khaled, and
  Richt{\'a}rik]{mishchenko2020random}
Konstantin Mishchenko, Ahmed Khaled, and Peter Richt{\'a}rik.
\newblock Random reshuffling: Simple analysis with vast improvements.
\newblock \emph{Advances in Neural Information Processing Systems},
  33:\penalty0 17309--17320, 2020.

\bibitem[Mitra et~al.(2021)Mitra, Jaafar, Pappas, and Hassani]{mitra2021linear}
Aritra Mitra, Rayana Jaafar, George~J Pappas, and Hamed Hassani.
\newblock Linear convergence in federated learning: Tackling client
  heterogeneity and sparse gradients.
\newblock \emph{Advances in Neural Information Processing Systems},
  34:\penalty0 14606--14619, 2021.

\bibitem[Nesterov(1983)]{nesterov1983:momentum}
Yurii Nesterov.
\newblock A method for solving the convex programming problem with convergence
  rate $o(1/k^2)$.
\newblock \emph{Proceedings of the USSR Academy of Sciences}, 269:\penalty0
  543--547, 1983.

\bibitem[Nguyen et~al.(2022)Nguyen, Malik, Zhan, Yousefpour, Rabbat, Malek, and
  Huba]{nguyen2022federated}
John Nguyen, Kshitiz Malik, Hongyuan Zhan, Ashkan Yousefpour, Mike Rabbat, Mani
  Malek, and Dzmitry Huba.
\newblock Federated learning with buffered asynchronous aggregation.
\newblock In \emph{International Conference on Artificial Intelligence and
  Statistics}, pages 3581--3607. PMLR, 2022.

\bibitem[Orvieto et~al.(2022{\natexlab{a}})Orvieto, Kersting, Proske, Bach, and
  Lucchi]{orvieto2022anticorrelated}
Antonio Orvieto, Hans Kersting, Frank Proske, Francis Bach, and Aurelien
  Lucchi.
\newblock Anticorrelated noise injection for improved generalization.
\newblock \emph{arXiv preprint arXiv:2202.02831}, 2022{\natexlab{a}}.

\bibitem[Orvieto et~al.(2022{\natexlab{b}})Orvieto, Raj, Kersting, and
  Bach]{orvieto2022explicit}
Antonio Orvieto, Anant Raj, Hans Kersting, and Francis Bach.
\newblock Explicit regularization in overparametrized models via noise
  injection.
\newblock \emph{arXiv preprint arXiv:2206.04613}, 2022{\natexlab{b}}.

\bibitem[Polyak(1964)]{polyak1964:momentum}
B.T. Polyak.
\newblock Some methods of speeding up the convergence of iteration methods.
\newblock \emph{USSR Computational Mathematics and Mathematical Physics},
  4\penalty0 (5):\penalty0 1--17, 1964.
\newblock ISSN 0041-5553.
\newblock \doi{https://doi.org/10.1016/0041-5553(64)90137-5}.
\newblock URL
  \url{https://www.sciencedirect.com/science/article/pii/0041555364901375}.

\bibitem[Robbins and Monro(1951)]{Robbins51:sgd_original}
Herbert Robbins and Sutton Monro.
\newblock A stochastic approximation method.
\newblock \emph{The Annals of Mathematical Statistics}, 22\penalty0
  (3):\penalty0 400 -- 407, 1951.
\newblock \doi{10.1214/aoms/1177729586}.
\newblock URL \url{https://doi.org/10.1214/aoms/1177729586}.

\bibitem[Shalev-Shwartz et~al.(2009)Shalev-Shwartz, Shamir, Srebro, and
  Sridharan]{ShalevShwartz2009:StochasticCO}
Shai Shalev-Shwartz, Ohad Shamir, Nathan Srebro, and Karthik Sridharan.
\newblock Stochastic convex optimization.
\newblock In \emph{Annual Conference Computational Learning Theory}, 2009.

\bibitem[Stich(2019)]{stich2018local}
Sebastian~U. Stich.
\newblock Local {SGD} converges fast and communicates little.
\newblock In \emph{International Conference on Learning Representations}, 2019.
\newblock URL \url{https://openreview.net/forum?id=S1g2JnRcFX}.

\bibitem[Stich and Karimireddy(2022)]{stich21:error-feedback}
Sebastian~U. Stich and Sai~Praneeth Karimireddy.
\newblock The error-feedback framework: Better rates for {SGD} with delayed
  gradients and compressed updates.
\newblock \emph{J. Mach. Learn. Res.}, 21\penalty0 (1), jun 2022.
\newblock ISSN 1532-4435.

\bibitem[Stich et~al.(2018)Stich, Cordonnier, and Jaggi]{stich2018sparsified}
Sebastian~U Stich, Jean-Baptiste Cordonnier, and Martin Jaggi.
\newblock Sparsified {SGD} with memory.
\newblock \emph{Advances in Neural Information Processing Systems}, 31, 2018.

\bibitem[Thakkar et~al.(2021)Thakkar, Andrew, and
  McMahan]{Thakkar19:adaptive_clipping}
Om~Thakkar, Galen Andrew, and H.~B. McMahan.
\newblock Differentially private learning with adaptive clipping.
\newblock In \emph{Advances in Neural Information Processing Systems}, 2021.

\bibitem[Tran and Cutkosky(2022)]{tran2022momentumFTRL}
Hoang Tran and Ashok Cutkosky.
\newblock Momentum aggregation for private non-convex erm, 2022.

\bibitem[Vardhan and Stich(2022)]{vardhan22:smoothing}
Harsh Vardhan and Sebastian~U. Stich.
\newblock Tackling benign nonconvexity with smoothing and stochastic gradients,
  2022.
\newblock URL \url{https://arxiv.org/abs/2202.09052}.

\bibitem[Wang et~al.(2017)Wang, Ye, and Xu]{wang2017differentially}
Di~Wang, Minwei Ye, and Jinhui Xu.
\newblock Differentially private empirical risk minimization revisited: Faster
  and more general.
\newblock \emph{Advances in Neural Information Processing Systems}, 30, 2017.

\bibitem[Wang et~al.(2019)Wang, Balle, and Kasiviswanathan]{wang2019subsampled}
Yu-Xiang Wang, Borja Balle, and Shiva~Prasad Kasiviswanathan.
\newblock Subsampled {R\'e}nyi differential privacy and analytical moments
  accountant.
\newblock In \emph{The 22nd International Conference on Artificial Intelligence
  and Statistics}, pages 1226--1235. PMLR, 2019.

\bibitem[Yu et~al.(2019)Yu, Jin, and Yang]{yu2019linear}
Hao Yu, Rong Jin, and Sen Yang.
\newblock On the linear speedup analysis of communication efficient momentum
  sgd for distributed non-convex optimization.
\newblock In \emph{International Conference on Machine Learning}, pages
  7184--7193. PMLR, 2019.

\bibitem[Yuan et~al.(2016)Yuan, Yang, Zhang, and Hao]{newton_step_mm}
Ganzhao Yuan, Yin Yang, Zhenjie Zhang, and Zhifeng Hao.
\newblock Convex optimization for linear query processing under approximate
  differential privacy, 2016.
\newblock URL \url{https://arxiv.org/abs/1602.04302}.

\bibitem[Yuan and Ma(2020)]{yuan2020federated}
Honglin Yuan and Tengyu Ma.
\newblock Federated accelerated stochastic gradient descent.
\newblock \emph{Advances in Neural Information Processing Systems},
  33:\penalty0 5332--5344, 2020.

\bibitem[Yun et~al.(2022)Yun, Rajput, and Sra]{yun2022minibatch}
Chulhee Yun, Shashank Rajput, and Suvrit Sra.
\newblock Minibatch vs local {SGD} with shuffling: Tight convergence bounds and
  beyond.
\newblock In \emph{International Conference on Learning Representations}, 2022.
\newblock URL \url{https://openreview.net/forum?id=LdlwbBP2mlq}.

\bibitem[Zhou et~al.(2019)Zhou, Liu, Li, Lin, Zhou, and Zhao]{zhou2019toward}
Mo~Zhou, Tianyi Liu, Yan Li, Dachao Lin, Enlu Zhou, and Tuo Zhao.
\newblock Toward understanding the importance of noise in training neural
  networks.
\newblock In \emph{International Conference on Machine Learning}, pages
  7594--7602. PMLR, 2019.

\bibitem[Zhu and Wang(2019)]{zhu2019poission}
Yuqing Zhu and Yu-Xiang Wang.
\newblock Poission subsampled {R\'e}nyi differential privacy.
\newblock In \emph{International Conference on Machine Learning}, pages
  7634--7642. PMLR, 2019.

\end{thebibliography}

\newpage
\appendix

\section{Additional Examples}\label{app:more_examples}
\subsection{Why the Frobenius Norm is not Predictive}
In this section we give an explicit example of a matrix $\mB$ for which the Frobenius norm $\norm{\mB}_F$ does not give a good estimation of the optimization behavior of \eqref{eq:opt-setup-matrix}.

\begin{example}[\chesspgd]\label{ex:chess-sgd} We consider the special case of algorithm \eqref{eq:opt-setup-matrix} whose noise correlation matrix $\mB$ whose lower triangle has a chess board-like structure given by
\begin{align*}
    \mB_{\chess} = \sqrt{2} \begin{pmatrix}
    1 & 0 & 0 & \dots & 0\\
    0 & 1 & 0 & \dots & 0\\
    1 & 0 & 1 & \dots & 0\\
    \dots \\
    0 & 1 & 0 & \dots & 1\\
    \end{pmatrix}
\end{align*}
\end{example}

\begin{figure}[ht]
    \centering
    \includegraphics[width=0.4\linewidth]{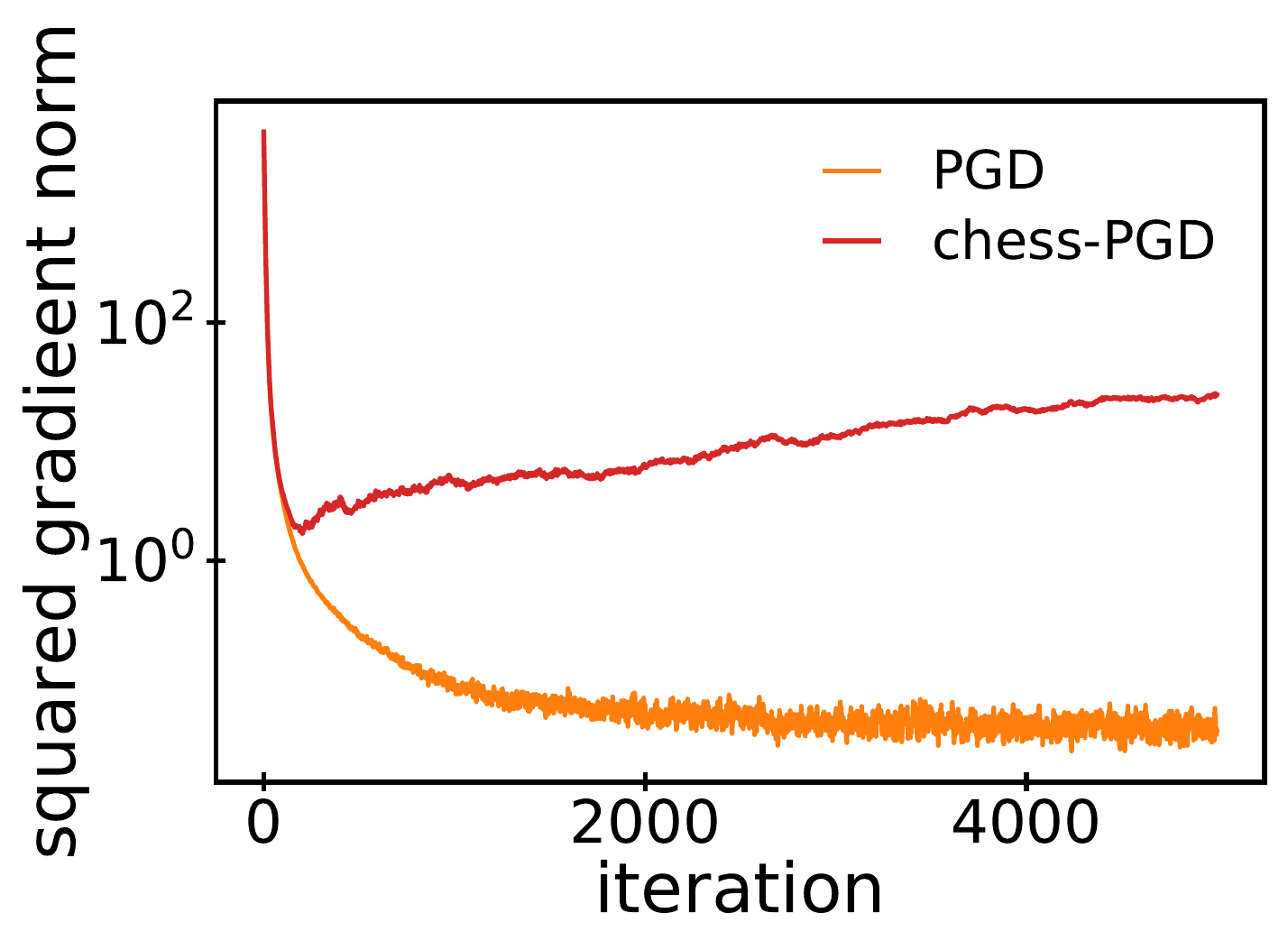}\hspace{4mm}
    \includegraphics[width=0.39\linewidth]{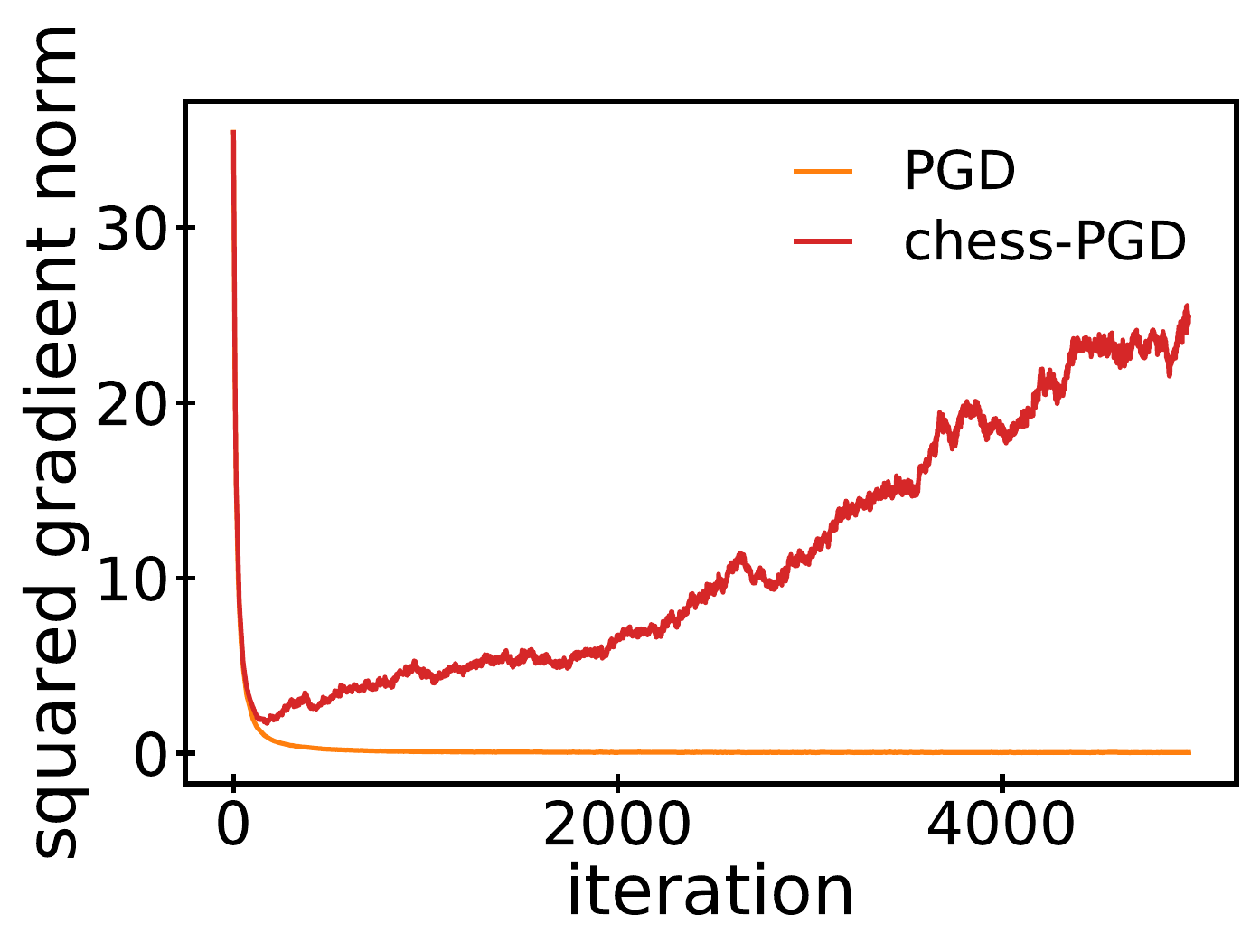}
    \caption{Comparison of \pgd and \chesspgd under the fixed stepsize, $\gamma = 0.02$. Y axis in the log scale on the left, and in the normal scale on the right.}
    \label{fig:PGD_vs_chess}
\end{figure}

We refer to this algorithm (whose perturbed noise structure is given by $\mB_{\chess}$) as \chesspgd.
Note that $\sens(\mC_{\chess})\norm{\mB_{\chess}}_F = \sens(\mC_{\mS}) \norm{\mS}_F$. Despite this, \pgd (for which $\mB = \mS$) converges strictly faster than \chesspgd in \cref{fig:PGD_vs_chess}. %

By contrast, our \cref{thm:main_convex} is better able to capture the behaviour of \chesspgd. Suppose that $\tau \leq \nicefrac{T}{4}$. Given a row $\bb_t$ of $\mB_{\chess}$, for any $t' < t$ we have
\[
\dfrac{t - t'}{2} \leq \norm{\bb_t - \bb_{t'}}^2 \leq t.
\]
Therefore, at least $\nicefrac{T}{4}$ of the summands in the noise term of \cref{thm:main_convex} are on the order of $\Theta(T)$. Plugging in this estimate into the convergence rate, we find that \chesspgd produces iterates that satisfy the convergence rate
\begin{equation}\label{eq:chess_sgd_rate}
\dfrac{1}{T+1}\sum_{t=0}^T \E\left[f(\xx_t) - f^*\right] = \tilde\cO\left(\dfrac{\norm{ \xx_0 - \xx^\star}^2}{\gamma T} + LT\gamma^2 \sigma^2 \right).
\end{equation}

Indeed, as we show below (and plot in Figure~\ref{fig:PGD_vs_chess}), \chesspgd linearly diverges with $T$ as predicted.

\subsection{Experimental Comparison of \pgd with \chesspgd}
In this section we illustrate that \chesspgd diverges while \pgd converges for the same quadratic functions as in \cref{sec:exp}. We set the stepsize constant, $\gamma = 0.02$. 
We plot $\|\nabla f(\xx_t)\|^2$ at each iteration $t$. We see that, as predicted by \eqref{eq:chess_sgd_rate}, \chesspgd diverges with linear rate in $T$, while \pgd converges to a constant noise level.

\section{Factorization Matrices}\label{app:weight_matrix}

As discussed in \Cref{sec:matrix-factorization}, \citet{denisov2022:matrix-fact} propose finding useful factorizations for DP training by solving the problem
\begin{equation}\label{eq:frob_objective_appendix}
\min_{\mB, \mC} \norm{\mB}_F^2~~\text{such that}~~\mB\mC = \mA,~\operatorname{sens}(\mC) = 1.
\end{equation}

As we discuss in \Cref{sec:better_factorization}, based on our convergence rates in \Cref{sec:theory}, we propose the following modified objective:

\begin{equation}\label{eq:frob_objective_modified}
\min_{\mB, \mC} \norm{\Lambda_\tau\mB}_F^2~~\text{such that}~~\mB\mC = \mA,~\operatorname{sens}(\mC) = 1.
\end{equation}

\begin{figure}[ht]
    \centering
    \includegraphics[width=0.35\linewidth]{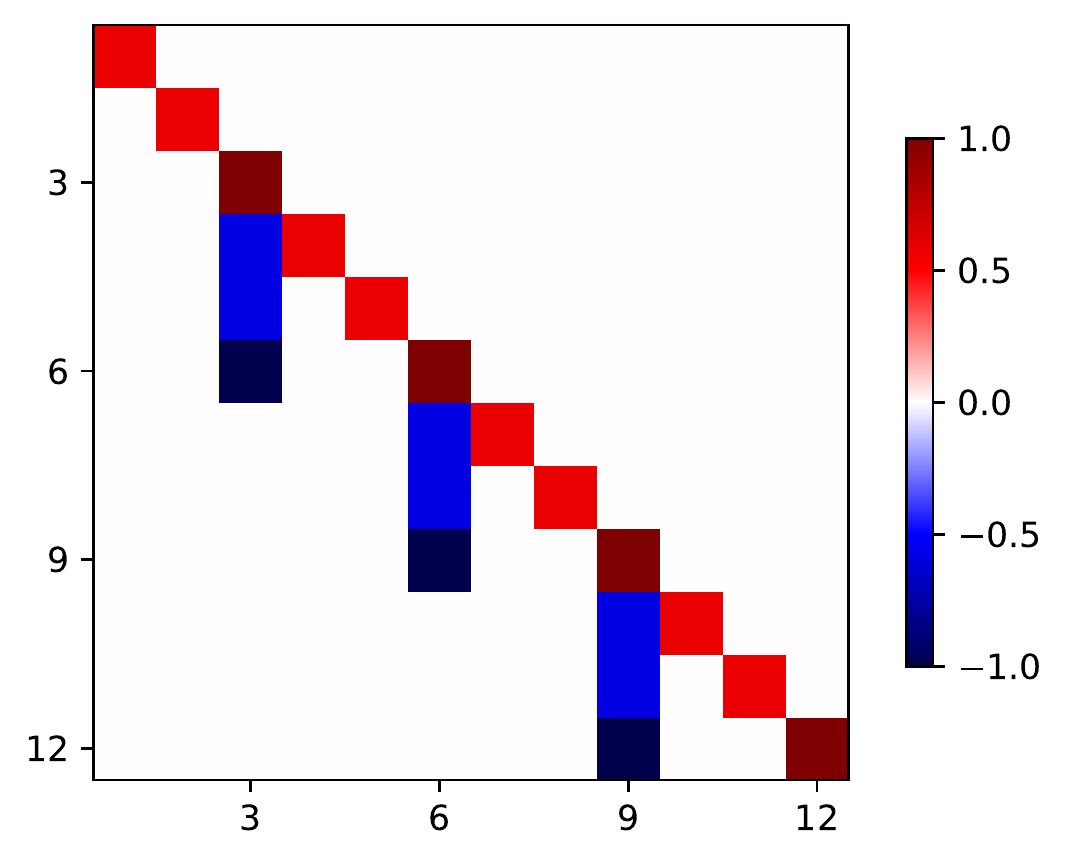}
    \caption{Elements of $\Lambda_{\tau}$ for $T = 12$, and $\tau = 3$.}
    \label{fig:lambda}
\end{figure}

The matrix $\Lambda_\tau = [\lambda_{tj}]_{t, j = 1, \dots, T}$ is defined as follows:
\begin{align*}
    \lambda_{tj} = \begin{cases}
    \frac{1}{\sqrt{\tau}} & j = t, \qquad ~t \neq 0  \bmod \tau \\
    - \frac{1}{\sqrt{\tau}} & j = \lfloor \frac{t}{\tau} \rfloor \tau, ~~ t \neq 0  \bmod \tau, t > \tau \\
    1 & j = t,\qquad ~t = 0 \bmod \tau\\
    - 1 & j = t - \tau,~~t = 0 \bmod \tau, t > \tau
    \end{cases}
\end{align*}
For all the other indices, $\lambda_{tj} = 0$. In Figure~\ref{fig:lambda} we give an example of such a matrix for $T = 12$ and $\tau = 3$. 

\begin{figure*}[h]
\centering     %
\subfigure[$\mB_{\text{MF}},~ T = 50$]{\label{fig:matrix_mf}\includegraphics[width=70mm]{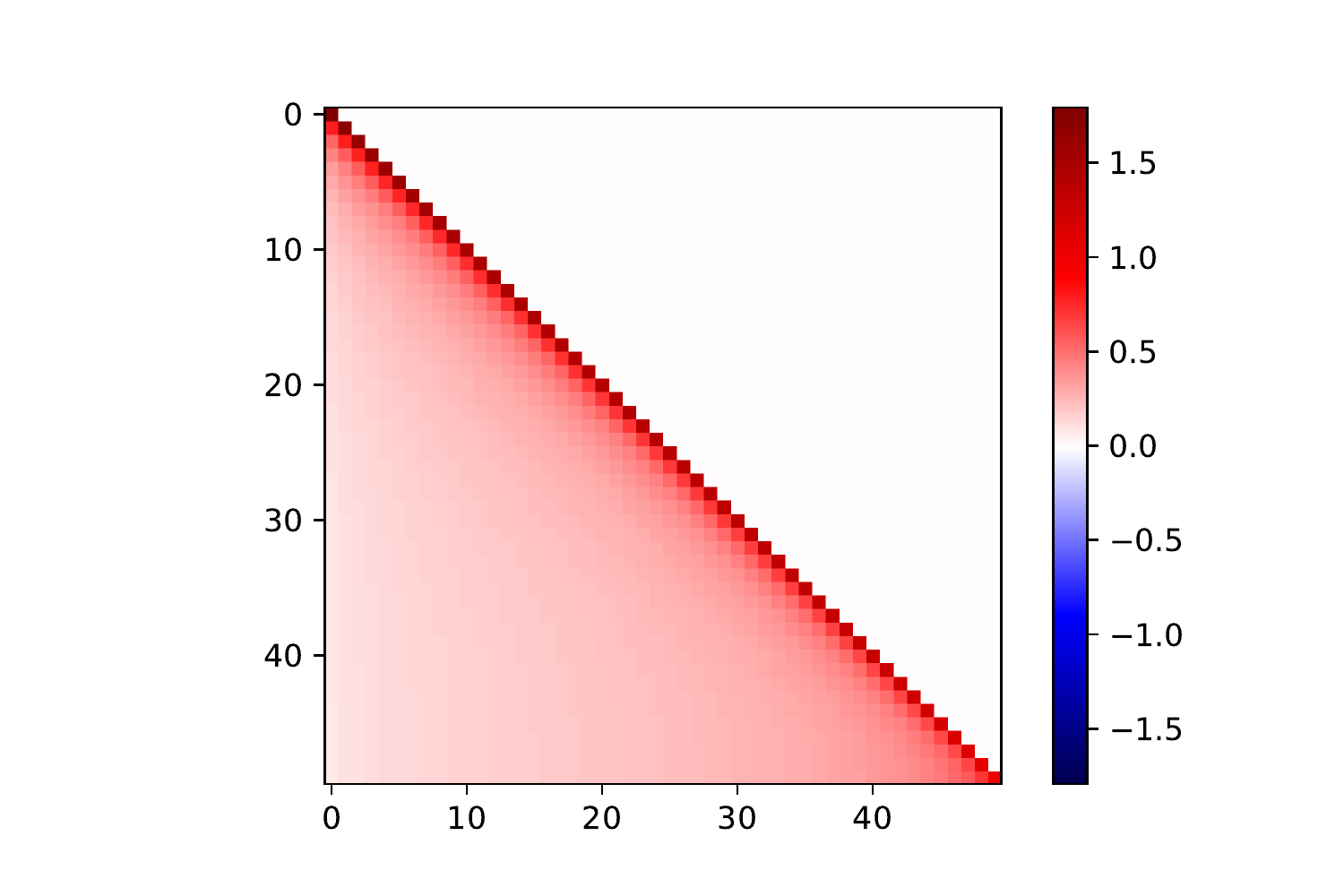}}%
\subfigure[$\mB_{\text{MF}^+},~ T = 50,~ \tau = 10$]{\label{fig:matrix_mfp}\includegraphics[width=70mm]{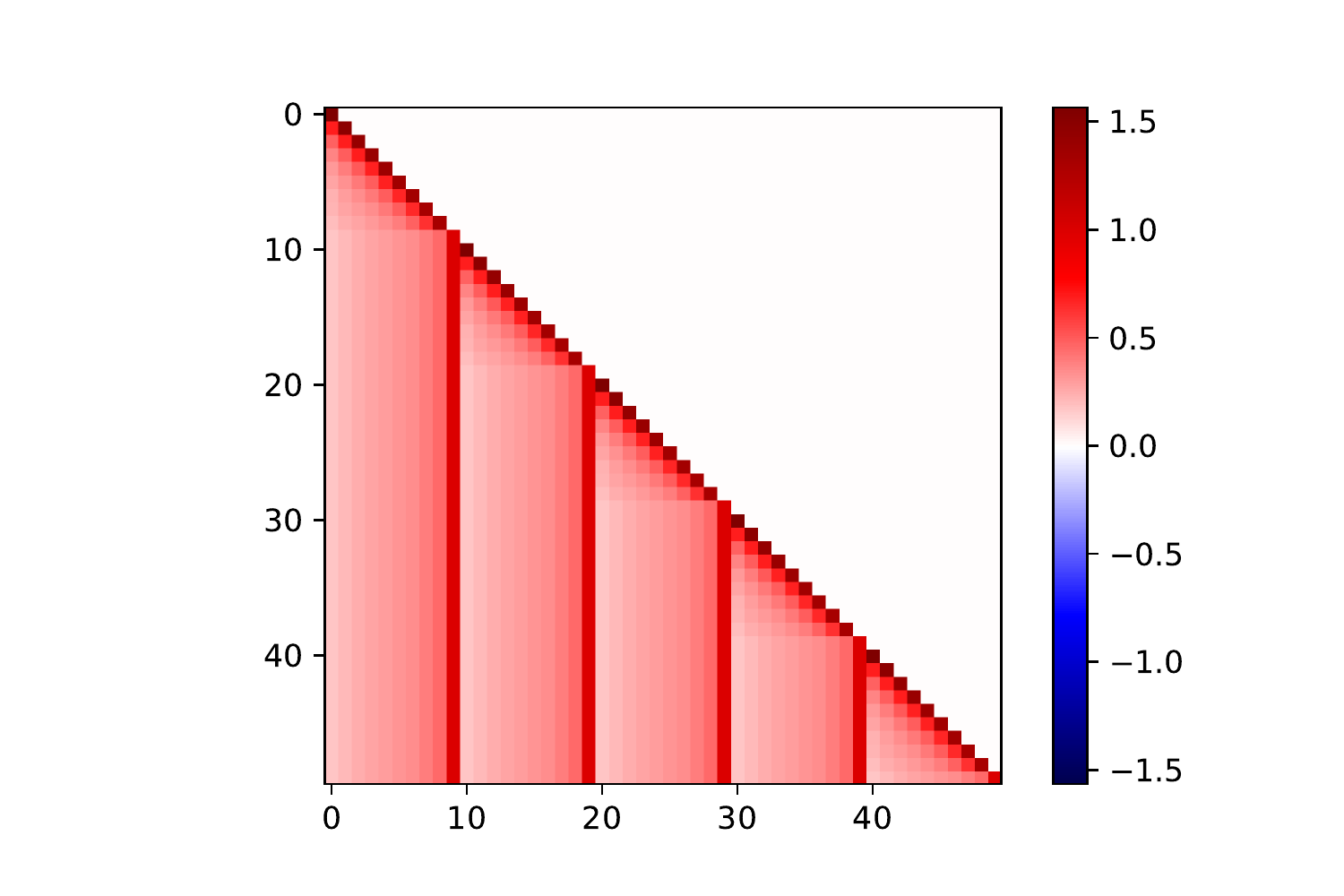}}\hspace{-2em}

\end{figure*}

To illustrate how the parameter $\tau$ affects the solution to the objective problem, we plot numerically computed approximate minimizers to \eqref{eq:frob_objective_appendix} and \eqref{eq:frob_objective_modified} in \Cref{fig:matrix_mf} and \Cref{fig:matrix_mfp}, respectively. Specifically, we plot the matrix $\mB$, and let $\mB_{\text{MF}}$ denote the solution to \eqref{eq:frob_objective_appendix} and $\mB_{\text{MF}^+}$ denote the solution to \eqref{eq:frob_objective_modified}. We can clearly see that for the latter, the parameter $\tau$ enforces a block-like structure such that the bands of correlation are at regular intervals of length $\tau$.%

\section{Proofs of Main Results}\label{app:main_proofs}
We analyse the algorithm with general $\mB$ that has the following iterates:
\begin{align}\label{eq:full_gd_with_matrix_noise}
    \begin{aligned}
    \xx_{t + 1} &= \xx_t - \eta ( \nabla f(\xx_t) + (\bb_{t + 1} - \bb_{t})^\top \mZ) & t \geq 1 
    \end{aligned}
\end{align}
where $\bb_{0} = 0$. We define $\vv_t = (\bb_{t + 1} - \bb_{t})^\top \mZ$ for $t \geq 0$, so that
\begin{align*}
    \xx_{t + 1} = \xx_t - \gamma \nabla f(\xx_t) - \gamma \vv_t
\end{align*}
For the analysis, we define a virtual sequence with restarts \eqref{eq:restart}, where we do restarts every $\tau$ iterations. Formally, we define virtual iterates $\{\tilde{\xx}_t\}_{t = 0}^T$ as follows:
\begin{align*}
    \tilde{\xx}_{t + 1} &= \tilde \xx_t - \gamma \nabla f(\xx_t) && \text{if }~t + 1 \neq 0 \bmod \tau\\
    \tilde \xx_{t + 1} &= \xx_{t + 1} && \text{if }~t + 1 = 0 \bmod \tau \text{  (restart iterations)}
\end{align*}
This means that $\tilde \xx_{k \tau} = \xx_{k \tau}$, for any nonnegative integer $k$.
\paragraph{Useful facts about this sequence.}
\begin{itemize}
    \item The closest restart iteration to $t$ is equal to $\lfloor \frac{t}{\tau}\rfloor \tau$.
    \item For $t < \tau$ we have 
    \begin{align*}
        \tilde \xx_t - \xx_t = \gamma \bb_{t}^\top \mZ
    \end{align*}
    \item For restart iterations $t + 1 = \tau$, 
    \begin{align*}
        \tilde \xx_{t + 1} - \xx_{t + 1} = 0
    \end{align*}
    \item For the next iteration just after restart $t + 1 = \tau + 1$
    \begin{align*}
        \tilde \xx_{\tau + 1} - \xx_{\tau + 1} = (\tilde \xx_{\tau} - \gamma \nabla f(\xx_\tau) ) - (\xx_\tau - \gamma \nabla f(\xx_\tau) -\gamma \vv_\tau) = \gamma \vv_\tau = \gamma (\bb_{\tau + 1} - \bb_{\tau})^\top \mZ 
    \end{align*}
    \item Thus, for arbitrary $t$,
    \begin{align}\label{eq:consensus_distance}
        \tilde \xx_{t} - \xx_{t} = \gamma (\bb_{t} - \bb_{\lfloor \frac{t}{\tau}\rfloor \tau})^\top \mZ
    \end{align}
    (and if $t = 0 \bmod \tau$, then the term cancels and we get $\tilde \xx_{t} - \xx_{t} = 0$), we assume that $\bb_{0} = \0$.
    \item We can re-write the restart iterations for $t + 1 = 0 \bmod \tau$
    \begin{align*}
        \tilde \xx_{t + 1} &= \xx_{t + 1} = \xx_t - \gamma \nabla f(\xx_t) - \gamma (\bb_{t + 1} - \bb_{t})^\top \mZ \\
        & = \tilde \xx_t - \gamma \nabla f(\xx_t) - \gamma (\bb_{t} - \bb_{\lfloor \frac{t}{\tau}\rfloor \tau})^\top \mZ - \gamma (\bb_{t + 1 } - \bb_{t})^\top \mZ \\
        & = \tilde\xx_t - \gamma\nabla f(\xx_t) - \gamma (\bb_{t + 1} -  \bb_{\lfloor \frac{t}{\tau}\rfloor \tau})^\top\mZ
    \end{align*}
    Equivalently, for $t + 1 = 0\bmod \tau$,
    \begin{align}\label{eq:restart-simplified}
        \tilde \xx_{t + 1} = \tilde \xx_{t} - \gamma \nabla f(\xx_{t}) - \gamma (\bb_{t + 1} - \bb_{t + 1 - \tau})^\top \mZ.
    \end{align}
\end{itemize}

\subsection{Assumptions and Useful Inequalities}
This section contains assumptions and inequalities that will be used throughout the proof. First, recall that in \cref{as:smooth}, we assume that $f$ is differentiable and $L$-smooth, so that
\begin{equation}\label{eq:L-smooth}
\forall \xx, \yy \in \R^d,~~~\norm{\nabla f(\xx) - \nabla f(\yy)} \leq L\norm{\xx - \yy}.
\end{equation}

In some settings, we will also assume convexity, so that
\begin{equation}\label{eq:convex}
\forall \xx, \yy \in \R^d,~~~f(\xx) - f(\yy) \leq \langle \nabla f(\xx), \xx-\yy\rangle.
\end{equation}

We will also make use of the following facts about the geometry of vectors in $\R^d$. 

\begin{lemma}
For any finite set of vectors $\{\aa_i\}_{i = 1}^{n} \subset \R^d$,
\begin{align}\label{eq:sum_of_n_vectors}
    \norm{\sum_{i = 1}^n \aa_i}^2 \leq n \sum_{i = 1}^n \norm{\aa_i}^2.
\end{align}
\end{lemma}
\begin{lemma}
For any two vectors $\aa, \bb \in \R^d$ and for all $\alpha > 0$,
\begin{align}\label{eq:scalar_product_ab}
    2 \langle \aa, \bb \rangle \leq \alpha \norm{\aa}^2 + \alpha^{-1}\norm{\bb}^2.
\end{align}
\end{lemma}

\subsection{Proof for Non-convex Functions}

\paragraph{Iterations without restarts.} If $t$ is such that $t \neq - 1 \bmod \tau$, where $k$ is some integer number, then between iteration $t$ and $t + 1$ no restart of virtual sequence happens and thus $\tilde \xx_{t  +1} = \tilde \xx_t - \gamma \nabla f(\xx_t)$. We follow closely standard perturbed iterate analysis \cite{mania17:perturbed_analysis, stich21:error-feedback}. By $L$-smoothness of $f$
\begin{align}
    f(\tilde \xx_{t + 1}) & \leq f(\tilde \xx_t) - \gamma \langle \nabla f(\tilde \xx_t), \nabla f(\xx_t) \rangle + \frac{L\gamma^2}{2} \norm{ \nabla f(\xx_t)}^2\nonumber \\
    & \leq f(\tilde \xx_t) - \frac{\gamma}{2} \norm{\nabla f(\tilde \xx_t)}^2 - \frac{\gamma}{2} \norm{\nabla f( \xx_t)}^2 + \frac{\gamma L^2}{2} \norm{\xx_t - \tilde \xx_t}^2 \nonumber \\
    & \stackrel{\eqref{eq:consensus_distance}}{\leq} f(\tilde \xx_t) - \frac{\gamma}{2} \norm{\nabla f(\tilde \xx_t)}^2 - \frac{\gamma}{2} \norm{\nabla f( \xx_t)}^2 + \frac{\gamma^3 L^2}{2} \norm{(\bb_{t} - \bb_{\lfloor \frac{t}{\tau}\rfloor\tau})^\top \mZ}^2\label{eq:descent-no-restart}
\end{align}
where on the second line we used that $- 2\langle \aa, \bb \rangle = - \norm{\aa}^2 - \norm{\bb}^2 + \norm{\aa - \bb}^2$ for any $\aa, \bb \in \R^d$.

\paragraph{Iterations with restarts.} Restart happens between iteration $t$ and $t + 1$ if $t = - 1 \bmod \tau$. In this case, the analysis is more involved. By $L$-smoothness and using update rule \eqref{eq:restart-simplified}
\begin{align}
    f(\tilde \xx_{t + 1}) &\leq f(\tilde \xx_t) - \gamma \langle \nabla f(\tilde \xx_t),  \nabla f(\xx_t) + (\bb_{t + 1} - \bb_{t + 1 - \tau})^\top \mZ\rangle \\
    & \qquad\qquad + \frac{L}{2} \gamma^2 \norm{ \nabla f(\xx_t) + (\bb_{t + 1} - \bb_{t + 1 - \tau})^\top \mZ}^2 \\
    &\stackrel{\eqref{eq:sum_of_n_vectors}}{\leq} f(\tilde \xx_t) - \underbrace{\gamma \langle \nabla f(\tilde \xx_t),  \nabla f(\xx_t) \rangle}_{:=T_1} - \underbrace{\gamma \langle \nabla f(\tilde \xx_t), (\bb_{t + 1} - \bb_{t + 1 - \tau})^\top \mZ \rangle}_{:=T_2} \\
    & \qquad\qquad + L \gamma^2 \norm{ \nabla f(\xx_t)}^2 + L \gamma^2 \norm{(\bb_{t + 1} - \bb_{t + 1 - \tau})^\top \mZ}^2 \label{eq:intermediate-descent}
\end{align}
We estimate separately the second and the third terms
\begin{align*}
    T_1 &= -\frac{\gamma}{2} \norm{\nabla f(\tilde \xx_t)}^2 -\frac{\gamma}{2} \norm{\nabla f(\xx_t)}^2 + \frac{\gamma}{2} \norm{\nabla f(\tilde \xx_t) - \nabla f(\xx_t)}^2 \\
    &\stackrel{\eqref{eq:L-smooth}}{\leq} -\frac{\gamma}{2} \norm{\nabla f(\tilde \xx_t)}^2 -\frac{\gamma}{2} \norm{\nabla f(\xx_t)}^2 + \frac{\gamma L^2}{2} \norm{\tilde \xx_t - \xx_t}^2\\
    &\stackrel{\eqref{eq:consensus_distance}}{\leq} -\frac{\gamma}{2} \norm{\nabla f(\tilde \xx_t)}^2 -\frac{\gamma}{2} \norm{\nabla f(\xx_t)}^2 + \frac{\gamma^3 L^2}{2} \norm{ (\bb_{t} - \bb_{\lfloor \frac{t}{\tau}\rfloor \tau})^\top \mZ}^2
\end{align*}
The third term,
\begin{align*}
    T_2 &= - \langle \nabla f(\tilde \xx_t), \gamma (\bb_{t + 1} - \bb_{t + 1 - \tau})^\top \mZ \rangle \\
    &\stackrel{\eqref{eq:scalar_product_ab},~\alpha = \frac{1}{8 L}}{\leq} \frac{1}{16 L}\norm{ \nabla f(\tilde \xx_t)}^2 + 4 L \gamma^2 \norm{(\bb_{t + 1} - \bb_{t + 1 - \tau})^\top \mZ}^2
\end{align*}
It is left to deal with the norm of the gradient $\frac{1}{16L}\norm{ \nabla f(\tilde \xx_t)}^2$. Using that $\tau = \frac{1}{L \gamma}$, and thus $\frac{1}{16L \tau} = \frac{\gamma}{16}$ we have
\begin{align*}
    \frac{1}{16 L}\norm{\nabla f(\tilde \xx_t)}^2  & = \frac{\gamma}{16} \sum_{i = 0}^{\tau - 1} \norm{ \nabla f(\tilde \xx_t)}^2 \\
    &\stackrel{\eqref{eq:sum_of_n_vectors}, \eqref{eq:L-smooth}}{\leq} \frac{\gamma}{8} \sum_{i = 0}^{\tau - 1} L^2 \norm{ \tilde \xx_t - \tilde \xx_{t - i}}^2 + \frac{\gamma}{8}\sum_{i = 0}^{\tau - 1} \norm{\nabla f(\tilde \xx_{t - i})}^2\\
    &\stackrel{\eqref{eq:descent-no-restart}}{\leq} \frac{\gamma}{8} \sum_{i = 1}^{\tau - 1} \gamma^2 L^2 \norm{ \sum_{j = t - i}^{t-1} \nabla f(\xx_{j})}^2 + \frac{\gamma}{8}\sum_{i = 0}^{\tau - 1} \norm{\nabla f(\tilde \xx_{t - i})}^2 \\
    &\stackrel{\eqref{eq:sum_of_n_vectors}}{\leq} \frac{\gamma^3 L^2}{8} \sum_{i = 1}^{\tau - 1} \tau \sum_{j = t - i}^{t-1} \norm{ \nabla f(\xx_{j})}^2 + \frac{\gamma}{8}\sum_{i = 0}^{\tau - 1} \norm{\nabla f(\tilde \xx_{t - i})}^2 \\
    & \leq \frac{\gamma^3L^2 \tau^2}{8} \sum_{i = 1}^{\tau - 1}\norm{ \nabla f(\xx_{t - i})}^2  + \frac{\gamma}{8}\sum_{i = 0}^{\tau - 1}\norm{\nabla f(\tilde \xx_{t - i})}^2\\
    & \stackrel{\tau = \frac{1}{\gamma L}}{\leq} \frac{\gamma}{8}\sum_{i = 1}^{\tau - 1}  \norm{ \nabla f(\xx_{t - i})}^2  + \frac{\gamma}{8}\sum_{i = 0}^{\tau - 1} \norm{\nabla f(\tilde \xx_{t - i})}^2
\end{align*}
Putting back our calculations of $T_1$ and $T_2$ into \eqref{eq:intermediate-descent}, and setting $\gamma \leq \frac{1}{4L}$ in order to estimate that $ L \gamma^2 \norm{ \nabla f(\xx_t)}^2 \leq \frac{\gamma}{4} \norm{\nabla f(\xx_t)}^2$
\begin{align}\label{eq:descent-restart}
    \begin{split}
    f(\tilde \xx_{t + 1}) &\leq f(\tilde \xx_t) - \frac{\gamma}{2} \norm{\nabla f(\tilde \xx_t)}^2 -\frac{\gamma}{4} \norm{\nabla f(\xx_t)}^2 + \frac{\gamma^3 L^2}{2} \norm{ (\bb_{t} - \bb_{\lfloor \frac{t}{\tau}\rfloor \tau})^\top\mZ}^2\\
    &\qquad\qquad +  5 L \gamma^2 \norm{(\bb_{t + 1} - \bb_{t + 1 - \tau})^\top \mZ}^2 + \frac{\gamma}{8}\sum_{i = 1}^{\tau - 1}  \norm{ \nabla f(\xx_{t - i})}^2  + \frac{\gamma}{8}\sum_{i = 0}^{\tau - 1} \norm{\nabla f(\tilde \xx_{t - i})}^2
    \end{split}
\end{align}

\paragraph{Combining iterations with and without restarts.}
It is left to average equations \eqref{eq:descent-no-restart} and \eqref{eq:descent-restart} over all iterations $0 \leq t \leq T$. We denote $\cT_1$ is the set of indices without restarts, and $\cT_2$ are restarts indices. 
\begin{align*}
    &\sum_{t \in \cT_1} \frac{\gamma}{8} \left(\norm{\nabla f(\tilde \xx_t)}^2 + \norm{\nabla f(\xx_t)}^2\right) + \sum_{t \in \cT_2} \frac{\gamma}{8}\left(\norm{\nabla f(\tilde \xx_t)}^2 + \norm{\nabla f(\xx_t)}^2 \right) \\
    &\qquad \qquad \leq (f(\xx_0) - f^\star) +  \frac{\gamma^3 L^2}{2}\sum_{t = 1}^T \norm{ (\bb_{t} - \bb_{\lfloor \frac{t}{\tau}\rfloor \tau})^\top \mZ}^2 + 5 L \gamma^2 \sum_{t \in \cT_1} \norm{(\bb_{t + 1} - \bb_{t + 1 - \tau})^\top \mZ}^2
\end{align*}
Dividing by $\frac{\gamma (T + 1)}{8}$, we get
\begin{align*}
    \frac{1}{T + 1}\sum_{t=0}^T \E\norm{\nabla f(\xx_t)}^2 \leq \frac{8 (f(\xx_0) - f^\star)}{\gamma (T + 1)} + \frac{4 \gamma^2 L^2}{T + 1}\sum_{t = 1}^T \E\norm{ (\bb_{t} - \bb_{\lfloor \frac{t}{\tau}\rfloor \tau})^\top \mZ}^2 \\
    + \frac{40 L \gamma}{T + 1} \sum_{k = 1}^{ \lfloor \frac{T}{\tau} \rfloor} \E\norm{(\bb_{k \tau} - \bb_{(k - 1) \tau})^\top \mZ}^2
\end{align*}
which completes the proof.

\subsection{Proof for Convex Functions}
Our proof for convex functions follows the same pattern as for non-convex: we consider separately iterations with and without restarts of the virtual sequence \eqref{eq:restart}. However, summing up these two cases is the most involved part of the proof in the convex case, and it is different from the non-convex case.

We will use the following fact in our proof.
\begin{lemma}
If function $f$ is convex \eqref{eq:convex}, $L$-smooth \eqref{eq:L-smooth}, and has a finite minimizer $x^*$, then
\begin{align}\label{eq:convex-smooth}
    \norm{\nabla f(\xx)}^2 \leq 2L \left( f(\xx) - f^\star \right).
\end{align}
\end{lemma}

\paragraph{Iterations without restarts.}
Using \eqref{eq:restart}, i.e. that $\tilde \xx_{t + 1} = \tilde \xx_t - \gamma \nabla f(\xx_t)$, for some point $\xx^\star$ that satisfies $\nabla f(\xx^\star) = 0$,
\begin{align*}
    \norm{\tilde\xx_{t + 1} - \xx^\star}^2 &= \norm{\tilde \xx_t - \xx^\star}^2 - 2\gamma \langle \nabla f(\xx_t),  \xx_t - \xx^\star \rangle + \gamma^2 \norm{\nabla f(\xx_t)}^2 + 2 \gamma \langle \nabla f(\xx_t), \xx_t - \tilde \xx_t \rangle \\
    &\stackrel{\eqref{eq:convex-smooth}, \eqref{eq:convex}}{\leq} \norm{\tilde \xx_t - \xx^\star}^2 - 2\gamma(1 - L \gamma) \left( f(\xx_t) - f^\star \right) + 2 \gamma \langle \nabla f(\xx_t), \xx_t - \tilde \xx_t \rangle
\end{align*}
We estimate the last term separately
\begin{align*}
    2 \langle \nabla f(\xx_t), \xx_t - \tilde\xx_t \rangle \stackrel{\eqref{eq:scalar_product_ab}, \alpha = 2L}{\leq} \frac{1}{2L} \norm{\nabla f(\xx_t)}^2 + 2L \norm{\xx_t - \tilde \xx_t}^2 \stackrel{\eqref{eq:convex-smooth}}{\leq} (f(\xx_t) - f^\star) + 2L \norm{\xx_t - \tilde \xx_t}^2
\end{align*}
Thus,
\begin{align}
    \norm{\tilde\xx_{t + 1} - \xx^\star}^2 &\leq \norm{\tilde\xx_t - \xx^\star}^2 - \gamma(1 - 2L \gamma)(f(\xx_t) - f^\star) + 2 L \gamma \norm{\xx_t - \tilde \xx_t}^2\nonumber\\
    &\stackrel{\gamma < \frac{1}{4L}, \eqref{eq:consensus_distance}}{\leq} \norm{\tilde\xx_t - \xx^\star}^2 - \frac{\gamma}{2}(f(\xx_t) - f^\star) + 2 L \gamma^3 \norm{(\bb_{t } - \bb_{\lfloor \frac{t}{\tau}\rfloor \tau })^\top \mZ}^2\label{eq:no-restart-convergence}
\end{align}

\paragraph{For the iterations with restarts. } This means that $t + 1 = k \tau$. Using \eqref{eq:restart-simplified},
\begin{align*}
    \norm{\tilde\xx_{t + 1} - \xx^\star}^2 &= \norm{\tilde \xx_{t} - \xx^\star - \gamma \nabla f(\xx_{t}) - \gamma (\bb_{t + 1} - \bb_{t + 1 - \tau})^\top \mZ }^2\\
    &= \norm{\tilde \xx_t - \xx^\star}^2 - 2 \gamma \langle \nabla f(\xx_{t}), \tilde \xx_{t} - \xx^\star\rangle - 2 \gamma \langle (\bb_{t + 1} - \bb_{t + 1 - \tau})^\top \mZ, \tilde \xx_{t} - \xx^\star\rangle \\
    & \qquad\qquad + \gamma^2 \norm{\nabla f(\xx_{t}) + (\bb_{t + 1} - \bb_{t + 1 - \tau})^\top \mZ}^2
\end{align*}
We estimate the second term same as in the case without restarts:
\begin{align*}
    - 2 \gamma \langle \nabla f(\xx_{t}), \tilde \xx_{t} - \xx^\star\rangle &= - 2 \gamma \langle \nabla f(\xx_{t}), \xx_{t} - \xx^\star\rangle - 2 \gamma \langle \nabla f(\xx_{t}), \tilde \xx_{t} - \xx_{t} \rangle\\
    &\leq - \gamma (f(\xx_t) - f^\star) + 2 L \gamma \norm{\xx_t - \tilde \xx_t}^2 
\end{align*}
For the last term,
\begin{align*}
    \gamma^2 \norm{\nabla f(\xx_{t}) + (\bb_{t + 1} - \bb_{t + 1 - \tau})^\top \mZ}^2 &\stackrel{\eqref{eq:sum_of_n_vectors}}{\leq} 2 \gamma^2 \norm{\nabla f(\xx_{t})}^2 + 2 \gamma^2 \norm{(\bb_{t + 1} - \bb_{t + 1 - \tau})^\top \mZ}^2\\
    &\stackrel{\eqref{eq:convex-smooth}}{\leq} 4 L \gamma^2 (f(\xx_t) - f^\star) + 2 \gamma^2 \norm{(\bb_{t + 1} - \bb_{t + 1 - \tau})^\top \mZ}^2
\end{align*}
Thus with $\gamma \leq \frac{1}{8 L}$,
\begin{align}\label{eq:restart-convergence}
\begin{split}
     \frac{\gamma}{2} (f(\xx_t) - f^\star) &\leq \norm{\tilde \xx_t - \xx^\star}^2 - \norm{\tilde\xx_{t + 1} - \xx^\star}^2 + 2 L \gamma^3 \norm{(\bb_{t} - \bb_{\lfloor \frac{t}{\tau}\rfloor \tau})^\top \mZ}^2 \\
     & \qquad \qquad + 2 \gamma^2 \norm{(\bb_{t + 1} - \bb_{t + 1 - \tau})^\top \mZ}^2 - 2 \gamma \langle (\bb_{t + 1} - \bb_{t + 1 - \tau})^\top \mZ, \tilde \xx_{t} - \xx^\star\rangle
\end{split}
\end{align}
\paragraph{Combining iterations with and without restarts.}
Summing up \eqref{eq:no-restart-convergence} and \eqref{eq:restart-convergence} for all $0 \leq  t \leq T$,
\begin{align}\label{eq:intermediate}
\begin{split}
    \frac{\gamma}{2}\sum_{t = 0}^{T} &(f(\xx_t) - f^\star) \leq \norm{\tilde \xx_0 - \xx^\star}^2 - \norm{\tilde\xx_{T + 1} - \xx^\star}^2 + 2 L \gamma^3 \sum_{t = 0}^{T} \norm{(\bb_{t} - \bb_{\lfloor \frac{t}{\tau}\rfloor \tau})^\top \mZ}^2\\
    & + 2 \gamma^2 \sum_{k = 1}^{ \lfloor \frac{T}{\tau} \rfloor} \E\norm{(\bb_{k \tau} - \bb_{(k - 1) \tau})^\top \mZ}^2 \underbrace{- 2\gamma \sum_{k = 1}^{\lfloor \frac{T}{\tau} \rfloor} \langle (\bb_{k \tau} - \bb_{(k - 1) \tau})^\top \mZ, \tilde \xx_{k \tau - 1} - \xx^\star\rangle}_{:=S_1}
\end{split}
\end{align}
We now separately estimate the last sum $S_1$. We first divide it in pairs of two consecutive terms, and sum each pair separately. Lets denote $t = k \tau - 1$ for some $k$. Sum of two consecutive terms with indexes $t$ and $t - \tau$ is equal to
\begin{align*}
    & - 2\gamma \langle (\bb_{t + 1} - \bb_{t + 1 - \tau})^\top \mZ, \tilde \xx_{t} - \xx^\star\rangle - 2\gamma \langle (\bb_{t + 1 - \tau} - \bb_{t + 1 - 2 \tau})^\top \mZ, \tilde \xx_{t - \tau} - \xx^\star\rangle \\
    = & - 2\gamma \langle (\bb_{t + 1} - \bb_{t + 1 - \tau})^\top \mZ, \tilde \xx_{t} - \xx^\star\rangle - 2\gamma \langle (\bb_{t + 1 - \tau} - \bb_{t + 1 - 2 \tau})^\top \mZ, \tilde \xx_{t} - \xx^\star\rangle \\
    &\qquad\qquad - 2\gamma \langle (\bb_{t + 1 - \tau} - \bb_{t + 1 - 2 \tau})^\top \mZ, \tilde \xx_{t - \tau} - \tilde \xx_{t} \rangle\\
    = & - 2\gamma \langle (\bb_{t + 1} - \bb_{t + 1 - 2 \tau})^\top \mZ, \tilde \xx_{t} - \xx^\star\rangle - 2\gamma \langle (\bb_{t + 1 - \tau} - \bb_{t + 1 - 2 \tau})^\top \mZ, \tilde \xx_{t - \tau} - \tilde \xx_{t} \rangle
\end{align*}
Using update rules \eqref{eq:restart}, it holds that $\tilde \xx_{t} = \tilde \xx_{t - \tau} - \gamma \sum_{j = t - \tau}^{t - 1}\nabla f(\xx_j) - \gamma (\bb_{t + 1 - \tau} - \bb_{t + 1 - 2\tau})^\top \mZ$, and thus
\begin{align*}
 - 2\gamma \langle  (\bb_{t + 1 - \tau}& - \bb_{t + 1 - 2 \tau})^\top \mZ, \tilde \xx_{t - \tau} - \tilde \xx_{t} \rangle \\
 &= - 2\gamma^2 \langle (\bb_{t + 1 - \tau} - \bb_{t + 1 - 2 \tau})^\top \mZ,  \sum_{j = t - \tau}^{t - 1}\nabla f(\xx_j) + (\bb_{t + 1 - \tau} - \bb_{t + 1 - 2\tau})^\top \mZ \rangle \\
 &= \sum_{j = t - \tau}^{t - 1} - 2\gamma^2 \langle (\bb_{t + 1 - \tau} - \bb_{t + 1 - 2 \tau})^\top \mZ,\nabla f(\xx_j) \rangle - 2 \gamma^2 \norm{ (\bb_{t + 1 - \tau} - \bb_{t + 1 - 2\tau})^\top \mZ}^2\\
 & \stackrel{\eqref{eq:scalar_product_ab}}{\leq} \gamma^2 \alpha \tau \norm{(\bb_{t + 1 - \tau} - \bb_{t + 1 - 2 \tau})^\top \mZ}^2 + \gamma^2 \alpha^{-1} \sum_{j = t - \tau}^{t - 1}\norm{\nabla f(\xx_t)}^2 \\
 & \qquad\qquad - 2 \gamma^2 \norm{(\bb_{t + 1 - \tau} - \bb_{t + 1 - 2\tau})^\top \mZ}^2\\
 & \stackrel{\alpha = \frac{2}{\tau}}{\leq} \frac{\gamma^2 \tau}{2} \sum_{j = t - \tau}^{t - 1}\norm{\nabla f(\xx_t)}^2
\end{align*}
Using these calculations, our original sum $S_1$ can be simplified as
\begin{align*}
    S_1 & \leq - 2\gamma \sum_{k = 1}^{\lfloor \frac{T}{2 \tau} \rfloor} \langle (\bb_{k \cdot 2 \tau} - \bb_{(k - 1) 2 \tau})^\top \mZ, \tilde \xx_{k \cdot 2 \tau - 1} - \xx^\star\rangle + \frac{\gamma^2 \tau }{2} \sum_{t = 0}^{\lfloor \frac{T}{\tau}\rfloor \tau - 2} \norm{\nabla f(\xx_t)}^2
\end{align*}
We reduced the sum of $\lfloor \frac{T}{\tau}\rfloor $ elements twice to the sum of the $\lfloor \frac{T}{2 \tau}\rfloor$ elements. Continuing in similar way, we will need to have $\log_2 \left(\lfloor \frac{T}{\tau}\rfloor\right)$ times until we reduce the original sum to just one element. Thus,
\begin{align*}
    S_1 &\leq - 2 \gamma \langle \bb_{\lfloor \frac{T}{\tau}\rfloor \tau }^\top \mZ , \tilde \xx_{\lfloor \frac{T}{\tau}\rfloor \tau} -\xx^\star \rangle + \frac{\gamma^2 \tau}{2} \log_2 \left(\left\lfloor \frac{T}{\tau}\right\rfloor\right) \sum_{t = 0}^{\lfloor \frac{T}{\tau}\rfloor \tau -2 } \norm{\nabla f(\xx_t)}^2\\
    &\stackrel{\eqref{eq:scalar_product_ab}, \alpha = 2}{\leq} \frac{1}{3}\norm{\tilde\xx_{\lfloor \frac{T}{\tau}\rfloor \tau } - \xx^\star}^2 +  3 \gamma^2 \norm{\bb_{\lfloor \frac{T}{\tau}\rfloor \tau}^\top \mZ}^2 + \frac{\gamma^2 \tau}{2} \log_2 \left(\left\lfloor \frac{T}{\tau}\right\rfloor\right) \sum_{t = 0}^{\lfloor\frac{T}{\tau}\rfloor\tau - 2} \norm{\nabla f(\xx_t)}^2
\end{align*}
We further transform the first term using the update rule \eqref{eq:restart}
\begin{align*}
 \tilde \xx_{T + 1}
 = \tilde \xx_{\lfloor \frac{T}{\tau}\rfloor\tau} - \gamma \sum_{j = \lfloor\frac{T}{\tau}\rfloor\tau}^T \nabla f(\xx_j)
 = \tilde \xx_{\lfloor \frac{T}{\tau}\rfloor\tau - 1} - \gamma \sum_{j = \lfloor\frac{T}{\tau}\rfloor\tau - 1}^T \nabla f(\xx_j) - \gamma \left(\bb_{\lfloor \frac{T}{\tau}\rfloor\tau } - \bb_{(\lfloor \frac{T}{\tau}\rfloor - 1)\tau }\right)^\top \mZ
\end{align*}
Thus, 
\begin{align*}
    \frac{1}{3}\norm{\tilde \xx_{\lfloor \frac{T}{\tau}\rfloor\tau} - \xx^\star}^2 \leq \norm{\tilde \xx_{T + 1} - \xx^\star}^2 + \gamma^2 \tau \sum_{j = \lfloor\frac{T}{\tau}\rfloor\tau - 1}^T \norm{\nabla f(\xx_j)}^2 + \gamma^2 \norm{\left(\bb_{\lfloor \frac{T}{\tau}\rfloor\tau} - \bb_{(\lfloor \frac{T}{\tau}\rfloor - 1)\tau}\right)^\top \mZ}^2
\end{align*}
And thus,
\begin{align*}
    S_1 &\leq \norm{\tilde\xx_{T + 1} - \xx^\star}^2 +  3 \gamma^2 \norm{\bb_{\lfloor \frac{T}{\tau}\rfloor \tau}^\top \mZ}^2 + \gamma^2 \tau \log_2 \left(\left\lfloor \frac{T}{\tau}\right\rfloor\right) \sum_{t = 0}^{T} \norm{\nabla f(\xx_t)}^2 \\
    & \qquad \qquad + \gamma^2 \norm{\left(\bb_{\lfloor \frac{T}{\tau}\rfloor\tau} - \bb_{(\lfloor \frac{T}{\tau}\rfloor - 1)\tau}\right)^\top \mZ}^2
\end{align*}

Choosing $\tau = \frac{1}{8 L \gamma \log_2(T)}$ ensures that $\gamma^2 \tau\log_2 \left(\left\lfloor \frac{T}{\tau}\right\rfloor\right) \leq \frac{\gamma}{8 L}$. Putting these calculations back into \eqref{eq:intermediate}, we get that
\begin{align*}
    \frac{\gamma}{2}\sum_{t = 0}^{T} (f(\xx_t) - f^\star) &\leq \norm{\tilde \xx_0 - \xx^\star}^2 + 3 L \gamma^3 \sum_{t = 0}^{T} \norm{(\bb_{t - 1} - \bb_{\lfloor \frac{t}{\tau}\rfloor \tau})^\top \mZ}^2 \\
    & \quad+ 3 \gamma^2 \sum_{k = 1}^{ \lfloor \frac{T}{\tau} \rfloor} \E\norm{(\bb_{k\tau} - \bb_{(k - 1) \tau })^\top \mZ}^2
     +\frac{\gamma}{8L}\sum_{t = 0}^{T} \norm{\nabla f(\xx_t)}^2 +  3 \gamma^2 \norm{\bb_{\lfloor \frac{T}{\tau}\rfloor\tau}^\top\mZ}^2
\end{align*}
Using \eqref{eq:convex-smooth}, we can further simplify 
\begin{align*}
    \frac{\gamma}{4}\sum_{t = 0}^{T} (f(\xx_t) - f^\star) &\leq 3 \gamma^2 \Big(L \gamma \sum_{t = 0}^{T} \norm{(\bb_{t} - \bb_{\lfloor \frac{t}{\tau}\rfloor \tau})^\top \mZ}^2 + \sum_{k = 1}^{ \lfloor \frac{T}{\tau} \rfloor} \E\norm{(\bb_{k\tau} - \bb_{(k - 1) \tau})^\top \mZ}^2 \\
    & \qquad \qquad + \norm{\bb_{\lfloor \frac{T}{\tau}\rfloor\tau}^\top\mZ}^2\Big) + \norm{\xx_0 - \xx^\star}^2
\end{align*}

\section{Convergence of \antipgd}\label{app:cancel}
Here we discuss the convergence of the Anti-PGD method, introduced in \cref{ex:cancel}.

Since $\tilde \xx_{t + 1} = \tilde \xx_t - \gamma \nabla f(\xx_t)$, for some point $\xx^\star$ that satisfies $\nabla f(\xx^\star) = 0$,
\begin{align*}
    \norm{\tilde\xx_{t + 1} - \xx^\star}^2 &= \norm{\tilde \xx_t - \xx^\star}^2 - 2\gamma \langle \nabla f(\xx_t),  \xx_t - \xx^\star \rangle + \gamma^2 \norm{\nabla f(\xx_t)}^2 + 2 \gamma \langle \nabla f(\xx_t), \xx_t - \tilde \xx_t \rangle \\
    &\stackrel{\eqref{eq:convex-smooth}, \eqref{eq:convex}}{\leq} \norm{\tilde \xx_t - \xx^\star}^2 - 2\gamma(1 - L \gamma) \left( f(\xx_t) - f^\star \right) + 2 \gamma \langle \nabla f(\xx_t), \xx_t - \tilde \xx_t \rangle
\end{align*}
We estimate the last term separately
\begin{align*}
    2 \langle \nabla f(\xx_t), \xx_t - \tilde\xx_t \rangle \stackrel{\eqref{eq:scalar_product_ab}, \alpha = 2L}{\leq} \frac{1}{2L} \norm{\nabla f(\xx_t)}^2 + 2L \norm{\xx_t - \tilde \xx_t}^2 \stackrel{\eqref{eq:convex-smooth}}{\leq} (f(\xx_t) - f^\star) + 2L \norm{\xx_t - \tilde \xx_t}^2
\end{align*}
Thus,
\begin{align*}
    \norm{\tilde\xx_{t + 1} - \xx^\star}^2 &\leq \norm{\tilde\xx_t - \xx^\star}^2 - \gamma(1 - 2L \gamma)(f(\xx_t) - f^\star) + 2 L \gamma \norm{\xx_t - \tilde \xx_t}^2\nonumber\\
    &\stackrel{\gamma < \frac{1}{4L}, \eqref{eq:consensus_distance}}{\leq} \norm{\tilde\xx_t - \xx^\star}^2 - \frac{\gamma}{2}(f(\xx_t) - f^\star)  + 2 L \gamma \norm{\xx_t - \tilde \xx_t}^2
\end{align*}

\section{Noise Lower Bound}\label{app:noise_lower_bound}
We consider function $f(\xx) = \frac{L}{2} \norm{\xx}^2$ that is convex and $L$-smooth, and we are running algorithm~\eqref{eq:opt-setup-vector} with constant stepsize $\gamma$, and we consider the two cases of $\mB = \mS$ and $\mB = \mI$.
\subsection{\pgd} This corresponds to Example~\ref{ex:pgd}. We will prove the lower bound on the noise term under the condition that $T$ is large enough, i.e. $T \geq \frac{\log 2}{\eta L}$.

Since $\nabla f(\xx) = L \xx$, the algorithm \eqref{eq:opt-setup-vector} takes a form
\begin{align*}
    \xx_{t + 1} = (1 - \gamma L ) \xx_t - \gamma \zz_{t + 1}
\end{align*}
Thus, since $\xx^\star = 0$,
\begin{align*}
    \E \norm{\xx_{t + 1}}^2  &= \E \norm{(1 - \gamma L ) \xx_t - \gamma \zz_{t + 1}}^2 = (1 - \gamma L)^2 \norm{\xx_t}^2  + \gamma^2 \sigma^2 \\
    &= (1 - \gamma L)^{2(t + 1)} \norm{\xx_0}^2  + \gamma^2 \sigma^2 \sum_{j = 0}^{t} (1 - \gamma L)^{2j}
\end{align*}
due to the unbiasedness and independence of $\zz_t$. We can exactly calculate the sum of this geometric series
\begin{align*}
    \sum_{j = 0}^{T - 1} (1 - \gamma L)^{2j} = \frac{1 - ( 1 - \gamma L )^{2T}}{1 - (1 - \gamma L)^2} = \frac{1 - ( 1 - 2 \gamma L )^{2T}}{ 2 \gamma L - \gamma^2 L^2} \geq \frac{1}{ 4 \gamma L}
\end{align*}
where at the last step we used that $ \gamma^2 L^2 > 0$ and that $T \geq \frac{\log 2}{\gamma L}$. 

And thus the function values are larger than
\begin{align*}
    f(\xx_T) - f^\star = \frac{L}{2} \norm{\xx_T}^2 \geq \frac{L}{2}(1 - \gamma L)^{2(t + 1)} \norm{\xx_0}^2  + \frac{1}{8} \gamma \sigma^2
\end{align*}
This shows that the noise term in \eqref{eq:sgd-rate} cannot be improved.

\subsection{\antipgd} This corresponds to Example~\ref{ex:cancel}. Since $\nabla f(\xx) = L \xx$, the algorithm \eqref{eq:opt-setup-vector} for \antipgd noise takes a form
\begin{align*}
    \xx_{t + 1} &= (1 - \gamma L) \xx_t - \gamma\zz_{t + 1} + \gamma\zz_{t} \\
    &= (1 - \gamma L)^2 \xx_{t - 1} - (1 - \gamma L) \gamma \zz_{t} + (1 - \gamma L) \gamma \zz_{t - 1} - \gamma \zz_{t + 1} + \gamma \zz_{t}\\
    &= (1 - \gamma L)^2 \xx_{t - 1} + (1 - \gamma L) \gamma \zz_{t - 1} - \gamma \zz_{t + 1} + \gamma^2 L \zz_{t}\\
    & = (1 - \gamma L)^{t + 1} \xx_{0} + (1 - \gamma L)^t \gamma \zz_1 + \gamma^2 L \sum_{j = 1}^{t - 1} (1 - \gamma L)^{t - j} \zz_{j + 1} - \gamma \zz_{t + 1}
\end{align*}
Thus,
\begin{align*}
    \E \norm{\xx_T}^2 &= ( 1- \gamma L )^{2T} \norm{\xx_0}^2 + (1 - \gamma L)^{2 (T - 1)} \gamma^2 \E \norm{ \zz_1}^2 + \gamma^4 L^2 \sum_{j = 1}^{T - 2} (1 - \gamma L)^{2 (T - 1 - j) } \E \norm{\zz_{j + 1}}^2 \\
    &+ \gamma^2 \E \norm{\zz_{t + 1}}^2 \geq ( 1- \gamma L)^{2T} \norm{\xx_0}^2 + \gamma^2 \sigma^2
\end{align*}
Thus the function values are larger than
\begin{align*}
    f(\xx_T) - f^\star = \frac{L}{2} \norm{\xx_T}^2 \geq ( 1- \gamma L)^{2T} \norm{\xx_0}^2 + \frac{L}{2}\gamma^2 \sigma^2
\end{align*}
This proves that the noise term in \eqref{eq:Anti-PGD-rate} cannot be improved.

\subsection{Virtual Sequence for \pgd}
In this section we show that for the \pgd algorithm, virtual sequences $\tilde \xx_t$ that are defined in \eqref{eq:virtual-standard} cannot give a tight convergence result. 

Since $\tilde \xx_{t + 1} = \tilde \xx_t - \gamma \nabla f(\xx_t)$, and $\nabla f(\xx_t) = L \xx_t$ we get
\begin{align*}
    \tilde \xx_{t + 1} = (1 - \gamma L) \tilde \xx_t + \gamma L (\tilde \xx_t - \xx_t) = (1 - \gamma L) \tilde \xx_t - \gamma^2 L \sum_{j = 0}^t \zz_j
\end{align*}
where the last equality is since $\tilde\xx_t - \xx_t = - \gamma\sum_{j = 1}^t\zz_j$.
Unrolling,
\begin{align*}
    \tilde \xx_{t + 1} = (1 - \gamma L) \tilde \xx_t + \gamma L (\tilde \xx_t - \xx_t) = (1 - \gamma L)^{t + 1} \tilde \xx_0 - \gamma^2 L \sum_{j = 1}^t \zz_j \sum_{i = 0}^j (1 - \gamma L)^i
\end{align*}
Thus the norm
\begin{align*}
    \E \norm{\tilde \xx_{T}}^2 = (1 - \gamma L)^{2 T} \norm{\xx_0}^2 + \gamma^4 L^2 \sum_{j = 1}^{T - 1} \left[\sum_{i = 0}^j (1 - \gamma L)^i\right]^2 \sigma^2
\end{align*}
We can calculate exactly the inner sum as
\begin{align*}
    \sum_{i = 0}^j (1 - \gamma L)^i = \frac{1 - (1 - \gamma L)^j}{\gamma L}
\end{align*}
and thus
\begin{align*}
    \E \norm{\tilde \xx_{T}}^2 = (1 - \gamma L)^{2 T} \norm{\xx_0}^2 + \gamma^2 \sum_{j = 1}^{T - 1} \left[1 - (1 - \gamma L)^j\right]^2 \sigma^2 \geq \gamma^2 \sum_{j = \frac{T}{2}}^{T - 1} \left[1 - 2 (1 - \gamma L)^j\right] \sigma^2
\end{align*}
It is left to note that for $T$ sufficiently large, $T \geq 2 \frac{\log 4}{\gamma L}$, it holds that $(1 - \gamma L)^{T/2} \leq \frac{1}{4}$ and thus $ \left[1 - 2 (1 - \gamma L)^j\right] \geq \frac{1}{2}$. Using this, we arrive
\begin{align*}
    \E \norm{\tilde \xx_{T}}^2  \geq \gamma^2 \sigma^2 \frac{T}{4}
\end{align*}
and this the function value $f(\tilde\xx_{T}) \geq L \gamma^2 \sigma^2 \frac{T}{8}$.

\section{Difficulties in Deriving a Unified Analysis}\label{app:difficulty}

In this section we explain the difficulties in unifying theoretical analysis using existing proof techniques described in the main text. In particular analysis through the real iterates $\xx_t$ can give good convergence guarantees only for \pgd, but not \antipgd, and vise versa, analysis through the virtual iterates $\tilde \xx_t$ can give a good convergence guarantee for \antipgd but not for \pgd. 

Directly analyzing \antipgd using the actual iterates $\xx_t$ of \eqref{eq:opt-setup-matrix}, we only get a convergence rate of
\begin{align*}
    \sum_{t = 0}^T \dfrac{\E \left[f(\xx_t) - f^\star\right]}{T+1} \leq \cO \left( \frac{\norm{ \xx_0 - \xx^\star}^2}{\gamma T} + \gamma \sigma^2  \right).
\end{align*}
Note that this is strictly worse than the \antipgd rate in \eqref{eq:Anti-PGD-rate}. While we do not see any fundamental limit to analysing \antipgd directly through its iterates $\xx_t$, we do not know of how to do so in a way that recovers the rate in \eqref{eq:Anti-PGD-rate}.

On the other hand, applying the perturbed iterate analysis (via the virtual sequence $\tilde \xx_t$ produced by \eqref{eq:opt-setup-matrix} when $\mZ = \0$) to \pgd, we only get a convergence rate of
\begin{align*}
    \sum_{t = 0}^T \dfrac{\E \left[f(\xx_t) - f^\star\right]}{T+1} \leq \cO \left( \frac{\norm{ \xx_0 - \xx^\star}^2}{\gamma T} + L T \gamma^2 \sigma^2  \right).
\end{align*}
This rate is strictly worse than the rate derived through a virtual sequence in \eqref{eq:sgd-rate} when $\gamma > \nicefrac{1}{L T}$. As we detail in Appendix~\ref{app:noise_lower_bound}, this bound is actually a tight upper bound for the convergence of the virtual sequence $f(\tilde \xx_t)$. However, the real sequence $\xx_t$ converges faster than this according to \eqref{eq:sgd-rate}. In short, while one can use the virtual sequence $\tilde\xx_t$ to effectively analyze anti-correlated noise, such techniques do not directly yield a tight analysis of \pgd.

\section{Applying Theorem~\ref{thm:main_convex} to special cases}
\paragraph{\pgd.} In this case, $\mB = \mS$ (\cref{ex:prefix-sum}), so if $i - j \leq \tau$ then $\norm{\bb_i - \bb_j}^2 \leq \tau$. The noise term in the convergence rate of Theorem~\ref{thm:main_convex} is therefore upper bounded by 
\begin{align*}
    \frac{\sigma^2 }{T L\tau}\Bigg[\frac{1}{\tau}\sum_{t = 1}^{T} \tau + \sum_{\substack{1 \leq t \leq T \\ t = 0\bmod{\tau}}} \tau + T \Bigg] = \tilde\cO\Bigg(\frac{\sigma^2}{L \tau}\Bigg) = \tilde\cO\Bigg(\gamma\sigma^2 \Bigg)
\end{align*}
This matches the tight convergence rate in \cref{prop:pgd}. 

\paragraph{\antipgd.} Since $\mB = \mI$, for any rows $\bb_i$, $\bb_j$, $\norm{\bb_i - \bb_j}^2 \leq 2$. Thus, the noise term in the convergence rate of Theorem~\ref{thm:main_convex} is upper bounded by
\begin{align*}
    \frac{\sigma^2 }{T L\tau}\Bigg[\frac{1}{\tau}\sum_{t = 1}^{T} 2 + \sum_{\substack{1 \leq t \leq T \\ t = 0\bmod{\tau}}} 2 + 1 \Bigg] = \tilde\cO\Bigg(\frac{\sigma^2}{L \tau^2}\Bigg) = \tilde\cO\Bigg(L \gamma^2\sigma^2 \Bigg)
\end{align*}
where we used $\tau = \tilde\cO(\nicefrac{1}{L\gamma})$. This recovers the tight convergence rate in \cref{prop:cancel}.

\section{Experiments}\label{app:more_experiments}
In this section we provide the complete experimental details for the experiments in Section \ref{sec:exp}, as well as additional experiments on the Stack Overflow dataset.
\subsection{Experiments with Quadratic Functions}

We study \emph{random quadratic} function $f(\xx) = \frac{1}{2}\norm{\mA \xx - \bb}^2$ to be able to precisely control the smoothness constant $L$ that appears in our theoretical analysis. In particular, we set the spectrum of $\mA \in \R^{100 \times 100}$ to have the values to be linearly distributed between $\lambda_{\min} = 0$ and $\lambda_{\max} = \sqrt{L}$, and we randomly shift the axis by unitary transformation. We calculate the unitary transformation by the SVD of a random matrix $\mD$ with every element $d_{ij} \in \cN(0, 1)$. Lets $\mD = \mU_D\mLambda_D\mV_D$ be the SVD decomposition, and let $\mLambda_A = \text{diag}(\lambda_{\max}, \dots, \lambda_{\min})$ is the matrix with the desired spectrum (between $\lambda_{\min} = 0$ and $\lambda_{\max} = \sqrt{L}$). We calculate the matrix $\mA$ as $\mA = \mU_D \mLambda_A \mV_D$. 
We also randomly sample the shift $\bb \in \cN(0, \mI)$, $\bb \in \R^{100}$.

We note that such quadratic function $f$ is $L$-smooth and convex. We fix the number of iterations $T$ to $5000$, and the variance of the noise $\sigma$ is equal to $20$.

In these experiments we aim to compare DP-MF, and our proposed DP-MF$^+$ methods under varying hyperparameter settings. We fix the smoothness $L = 10$, 
and we vary the learning rate $\gamma$ over the logarithmic grid between $10^{-4}$ and $1$, and we further select the region of learning rates around the optimal $\gamma$. We also tune parameter $\tau$ in DP-MF$^+$ over the grid $\{1, 2, 10, 50, 100, 200, 500, 1000, 5000\}$. 

\subsection{Practical DP Training Experiments}

\begin{table}
    \centering
    \begin{adjustbox}{width=0.6\linewidth}
    \begin{tabular}{c|ccc}
        Dataset & MNIST & CIFAR-10 & StackOverflow \\
        \midrule 
        Train Records & 60,000 & 50,000 & 135,818,730 \\
        Test Records & 10,000 & 10,000 & 16,586,035 \\
        Dimensionality & 784 & 3,072 & 200,000 \\
        Classes & 10 & 10 & 10,000 \\
        Model & Logistic & CNN & LSTM \\
        Privacy Unit & Example & Example & User \\
        Parameters & 7,056 & 550,570 & 4,050,748 \\
        Learning Setting & Centralized & Centralized & Federated \\
    \end{tabular}
    \end{adjustbox}
    \vspace{2mm}
    \caption{Summary of datasets and associated problems considered in this empirical evaluation.}
    \label{tab:exp_summary}
\end{table}

\paragraph*{Datasets and tasks.}
\cref{tab:exp_summary} summarizes the datasets and problems used in our empirical evaluation.  
For the MNIST dataset, light preprocessing is done so the $28 \times 28$ input images are flattened to size $784$ vectors and normalized so entries lie in the range $[0,1]$.  For the CIFAR-10 and Stack Overflow datasets, the experimental setup including data preprocessing follows exactly from \citet{denisov2022:matrix-fact} and \citet{choquette22:multi-epochs}.

\paragraph*{Metrics.}

For each dataset, mechanism, and privacy parameter, we run the mechanism for multiple trials and report the test set accuracy of the final iterate.  We compute the mean and standard error of the reported test set accuracies.

\paragraph*{MNIST, logistic regression.}

For MNIST we train a logistic regression model to predict image labels. All mechanisms train for $T = 2048$ iterations and either $1$ or $16$ epochs, corresponding to batch sizes of 29 and 469 respectively.\footnote{In practice, one often trains small-scale models for many epochs, perhaps even using full-batch gradients, to improve the privacy/utility trade-off (at the cost of increased computation). We are interested in the \emph{relative} performance for a fixed computation budget, so we train for a small number of epochs.} We vary $\epsilon$ over $\{0.01, 0.1, \dots, 100\}$ and fix $\delta = 10^{-6}$.  We fix the clipping threshold at $1.0$ and the learning rate at $0.5$. We run each experiment for $5$ trials, and plot the mean test set accuracy along with error bars indicating the standard error of the estimate.  

\paragraph*{CIFAR-10, CNN.}

For CIFAR-10, we follow the experimental setup from \citep{choquette22:multi-epochs} and train a CNN model to predict image labels. Specifically, we train all mechanisms for $20$ epochs and $T = 2000$ iterations, which corresponds to a batch size of $500$.\footnote{While \citet{choquette22:multi-epochs} use momentum and learning rate decay, we omit the use of such techniques as they are orthogonal to our theoretical results.}   We consider $\epsilon = 1, 2, 4, 8, 16, 32$ and set $\delta = 10^{-6}$.  We tune the learning rate non-privately for each method and $\epsilon$ by running a single trial with a fixed random seed and choosing the one which achieved the lowest training error.  For each value of $\epsilon$, we use the tuned learning rate and run $12$ new trials with different random seeds, and record the test set accuracy at the end of training. 

\paragraph{Stack Overflow, LSTM.} We follow the experimental setup of \citet{denisov2022:matrix-fact}, and train a next-word prediction LSTM model on the Stack Overflow dataset~\citep{stackoverflow}. We train each mechanism for $1$ epoch and $2048$ iterations, which corresponds to about $167$ clients per round, each holding an average of $\approx 400$ records.  We vary the hyper-parameters according to prior work and run $2$ trials for each hyper-parameter setting. We report results for the best hyper-parameters setting of each mechanism. We use federated averaging instead of gradient descent. Additionally, to be consistent with the prior work and to test if our proposed factorizations are compatible with the other types of workloads, we use momentum and learning rate decay. Although the $\mC$ matrix was optimized for the Prefix workload, $\mA = \mS$, it is applied to a variant $\mA=\mS'$ that incorporates momentum and learning rate decay by setting $\mB = \mS' \mC^{-1}$.  
More details of how DP-MF and DP-MF$^+$ apply to this setting are available in \citet{denisov2022:matrix-fact}. 

The results are shown in \cref{tab:stackoverflow} for varying the noise multiplier, which corresponds to values of $\epsilon$ are equal to $\{17.65, 7.6, 3.44, 1.61, 0.76\}$. We see no significant difference between DP-MF and DP-MF$^+$, as the small differences in performance are within the statistical bounds one would expect if they had identical means. At larger noise multipliers, both DP-MF and DP-MF$^+$ exhibit learning instabilities.

\begin{table}[t]
    \centering
\begin{tabular}{c|cc}
\toprule
Noise Multiplier & DP-MF & DP-MF$^+(\tau=2048)$ \\\hline
\midrule
0.341            &                            $ 24.63 \pm 0.06 $ &                                 $ 24.58 \pm 0.12 $ \\
0.682            &                            $ 23.76 \pm 0.14 $ &                                 $ 23.73 \pm 0.16 $ \\
1.364            &                            $ 22.54 \pm 0.11 $ &                                 $ 22.44 \pm 0.08 $ \\
2.728            &                           $ 11.51 \pm 12.71 $ &                                $ 10.42 \pm 13.05 $ \\
5.456            &                             $ 0.03 \pm 0.02 $ &                                  $ 0.05 \pm 0.06 $ \\
\bottomrule
\end{tabular}
    \vspace{2mm}
    \caption{\label{tab:stackoverflow} Comparison of test set accuracies on the Stack Overflow next word prediction task between DP-MF and DP-MF$^+$.}
\end{table}

\clearpage

\end{document}